\documentclass[manuscript, screen]{acmart}

\AtBeginDocument{%
  \providecommand\BibTeX{{%
    \normalfont B\kern-0.5em{\scshape i\kern-0.25em b}\kern-0.8em\TeX}}}

\setcopyright{acmlicensed}
\copyrightyear{2024}
\acmYear{2024}
\acmDOI{XXXXXXX.XXXXXXX}

\usepackage{amssymb}

\usepackage{pifont}

\usepackage{multirow}

\usepackage{enumitem}
\usepackage{bm}

\usepackage{myextarrows}

\definecolor{attn_light_pink}{RGB}{240, 201, 201}
\definecolor{attn_pink}{RGB}{239, 148, 158}
\definecolor{memory_light_blue}{RGB}{181, 209, 234}
\definecolor{memory_blue}{RGB}{46, 189, 241}
\definecolor{pe_light_orange}{RGB}{246, 205, 173}
\definecolor{pe_orange}{RGB}{249, 180, 145}
\definecolor{context_light_green}{RGB}{156, 225, 219}
\definecolor{context_green}{RGB}{48, 192, 180}
\definecolor{mis_purple}{RGB}{220, 205, 231}
\definecolor{block_purple}{RGB}{192, 164, 213}
\definecolor{loss_purple}{RGB}{112, 48, 160}

\definecolor{checkmarkcolor}{RGB}{255, 105, 105} 
\definecolor{xmarkcolor}{RGB}{220, 220, 220}

\newcommand{\defeq}{\mathrel{\mathop:}=}
\newcommand{\transpose}{^{\mathrm{T}}}
\newcommand{\domain}[1]{\mathbb{R}^{#1}}
\newcommand{\sminus}{\text{-}}
\newcommand{\xmark}{\ding{55}}
\newcommand{\figlabel}[1]{\textbf{\textit{(#1)}}}

\newcommand{\coloredcheckmark}{\textcolor{checkmarkcolor}{\checkmark}}
\newcommand{\coloredxmark}{\textcolor{xmarkcolor}{\xmark}}

\emergencystretch=\maxdimen
\begin{document}



\title{
    Advancing Transformer Architecture in Long-Context Large Language Models: A Comprehensive Survey
}



\author{Yunpeng Huang}
\email{hyp@smail.nju.edu.cn}
\affiliation{%
  \institution{State Key Lab of Novel Software Technology, Nanjing University}
  \country{China}
  \postcode{210023}
}

\author{Jingwei Xu}\authornote{Corresponding author.}
\email{jingweix@nju.edu.cn}
\affiliation{%
  \institution{State Key Lab of Novel Software Technology, Nanjing University}
  \country{China}
  \postcode{210023}
}

\author{Junyu Lai}
\email{junyu\_lai@smail.nju.edu.cn}
\affiliation{%
  \institution{State Key Lab of Novel Software Technology, Nanjing University}
  \country{China}
  \postcode{210023}
}

\author{Zixu Jiang}
\email{jzxthlw@gmail.com}
\affiliation{%
  \institution{State Key Lab of Novel Software Technology, Nanjing University}
  \country{China}
  \postcode{210023}
}

\author{Taolue Chen}
\email{t.chen@bbk.ac.uk}
\affiliation{%
  \institution{School of Computing and Mathematical Sciences, Birkbeck, University of London}
  \city{London}
  \country{UK}
}

\author{Zenan Li}
\email{lizn@smail.nju.edu.cn}
\affiliation{%
  \institution{State Key Lab of Novel Software Technology, Nanjing University}
  \country{China}
  \postcode{210023}
}

\author{Yuan Yao}
\email{y.yao@nju.edu.cn}
\affiliation{%
  \institution{State Key Lab of Novel Software Technology, Nanjing University}
  \country{China}
  \postcode{210023}
}

\author{Xiaoxing Ma}
\email{xxm@nju.edu.cn}
\affiliation{%
  \institution{State Key Lab of Novel Software Technology, Nanjing University}
  \country{China}
  \postcode{210023}
}

\author{Lijuan Yang}
\email{yanglijuan04@baidu.com}

\author{Hao Chen}
\email{chenhao52@baidu.com}

\author{Shupeng Li}
\email{lishupeng@baidu.com}

\author{Penghao Zhao}
\email{zhaopenghao@baidu.com}
\affiliation{%
  \institution{Baidu.inc}
  \city{Beijing}
  \country{China}
}

\renewcommand{\shortauthors}{Huang, et al.}



\begin{abstract}
Transformer-based Large Language Models (LLMs) have been 
applied in diverse areas such as knowledge bases, human interfaces, and dynamic agents, and
marking a stride towards achieving Artificial General Intelligence (AGI).
However, current LLMs are predominantly pretrained on short text snippets, which compromises their effectiveness in processing the long-context prompts that are frequently encountered in practical scenarios.
This article offers a comprehensive survey of the recent advancement in Transformer-based LLM architectures aimed at enhancing the long-context capabilities of LLMs throughout the entire model lifecycle, from pre-training through to inference.
We first delineate and analyze the problems of handling long-context input and output with the current Transformer-based models. We then provide a 
taxonomy 
and the landscape of upgrades on Transformer architecture to solve these problems. 
Afterwards, we provide an investigation on wildly used evaluation necessities tailored for long-context LLMs, including datasets, metrics, and baseline models, as well as optimization toolkits such as libraries, frameworks, and compilers to boost the efficacy of LLMs across different stages in runtime. Finally, we discuss the 
challenges and potential avenues for future research. 
%
A curated repository of relevant literature, continuously updated, is available
at~\url{https://github.com/Strivin0311/long-llms-learning}.
\end{abstract}

\ccsdesc[500]{Computing methodologies~Neural networks}
\ccsdesc[500]{General and reference~Surveys and overviews}
\ccsdesc[500]{Computing methodologies~Natural language processing}
\ccsdesc[500]{Computer systems organization~Neural networks}
\ccsdesc[500]{General and reference~Evaluation}
\ccsdesc[500]{Software and its engineering~Software libraries and repositories}
\ccsdesc[500]{Computing methodologies~Parallel algorithms}
\ccsdesc[500]{Software and its engineering~Memory management}



\keywords{
    large language models,
    long context,
    Transformer architecture, 
    deep learning
}


\received{XX XXXX 20XX}
\received[revised]{XX XXXX 20XX}
\received[accepted]{XX XXXX 20XX}


\maketitle


    \section{Introduction}\label{sec:introduction}

In recent years, 
fueled by Transformer-based models such as BERT~\cite{devlin2018bert}, 
GPT~\cite{radford2018improving, radford2019language, brown2020language} and their variants~\cite{raffel2020exploring, touvron2023llama2, jiang2023mistral}, Natural Language Processing (NLP) has seen significantly advancement 
in human language understanding and generation~\cite{therasa2022survey, li2022pretrained}, revolutionizing numerous tasks in Natural Language Understanding (NLU) such as sentiment analysis~\cite{zhang2018deep}, Natural Language Generation (NLG) such as document summarization~\cite{el2021automatic}, as well as other domains such as computer vision~\cite{khan2022transformers} and autonomous driving~\cite{hu2023planning}. 
In particular, in the wake of ChatGPT~\cite{openai-blog-chatgpt}, PaLM2~\cite{anil2023palm}, GPT4~\cite{openai-blog-gpt4, openai2023gpt4}, Claude2~\cite{modelcard2023claude2}, etc, the Transformer-based Large Language Models (LLMs) which scale up to 1B---100B 
parameters to empower emergence abilities~\cite{wei2022emergent} have shown a new exhilarating path towards Artificial General Intelligence (AGI)~\cite{bubeck2023sparks}, and have been rapidly adopted in a myriad of human-interactive applications such as chatbots~\cite{rudolph2023war, lee2023prompted}, programming assistants~\cite{wermelinger2023using, yeticstiren2023evaluating} and educational tutors~\cite{abd2023large, milano2023large}. 


Transformer is an intricate deep neural network model, which integrates several preceding designs~\cite{bahdanau2014neural, he2016deep, ba2016layer} and novel components to support sequence-to-sequence language modeling, initially in machine translation~\cite{vaswani2017attention}. Contemporary LLMs largely adopt Transformer architecture, leveraging its modules~\cite{devlin2018bert, radford2018improving, raffel2020exploring}, among which they own the success mainly due to their well-designed attention mechanism that captures global dependencies of each pair of tokens across the whole input, enabling the model to handle sequences with intricate relations. However, its quadratic time and space complexities pose significant computational resource challenges, limiting input text length during training and effective context window during inference. Additionally, the lack of a robust and generalizable mechanism for positional embeddings (PEs) leads to performance degradation and fluctuation during inference, particularly with longer sequences or position shifting on relevant information~\cite{liu2024lost}.

With LLMs deeply ingrained in various applications that require long-context comprehension~\cite{yang2018hotpotqa, kočiský2017narrativeqa} and generation~\cite{lu2021codexglue, huang2021efficient}, the demand for long-context LLMs capable of comprehending and generating extremely long sequences effectively and efficiently becomes increasingly indispensable and urgent. 
Consequently, researchers have devoted significant efforts to enhancing the Transformer architecture to address the long-context problem in LLMs, including optimization on the efficiency of attention (Section \ref{subsec: effcient_attention}), context window extension with extra memory mechanisms (Section \ref{subsec: long-term_memory}), effective length generalization with extrapolative PEs (Section \ref{subsec: extrapolative_pe}), context pre/postprocessing (Section \ref{subsec: context_processing}), and other miscellaneous methods (Section \ref{subsec: miscellaneous}) such as specific pretraining objectives, mixture of experts (MoE), quantization, parallelism, etc.

\smallskip
\noindent\textbf{Existing surveys.} The field of long-context LLMs has become one of 
the most rapidly developing research areas on LLMs recently, with some existing surveys~\cite{koh2022empirical, tay2022efficient, lin2022survey, long-context-transformers, dong2023survey}. 
\cite{koh2022empirical} offers an overview of long document summarization, but does not delve into 
techniques of long text modeling.~\cite{tay2022efficient} and ~\cite{lin2022survey} primarily concentrate on improving the computational efficiency of Transformers in long-text scenarios. Although~\cite{long-context-transformers} underscores the challenges LLMs face when engaging with extensive sequences, its discussed methods predominantly align with efficient Transformers, similar to~\cite{tay2022efficient} and~\cite{lin2022survey}. 
A more recent survey~\cite{dong2023survey} bears the closest resemblance to our study, 
but is considerably less comprehensive 
than ours. In particular, we review the advancement in breaking the barriers of context length across all stages for more intricate and scalable Transformer-based LLMs by exploring the Transformer
from both an algorithmic design and system architecture perspective.

This survey aims to present 
a panorama of literature on architecture evolution for scaling the effective context window length of the state-of-the-art Transformer-based LLMs.
The main contributions are as follows.

\begin{itemize}
    \item We provide a holistic taxonomy 
    by breaking down the Transformer architecture and then delving into the existing methods in enhancing long-context LLMs during stages including pretraining, fine-tuning, inference and pre/postprocessing.
    \item We explore the widely-used evaluation necessities, comprising datasets, metrics, and baseline specifically assessing the long-context capabilities of LLMs, followed by some popular toolkits to optimize LLMs' efficiency and effectiveness for both training and inference, such as libraries, frameworks, and compilers.
    \item We identify key challenges to revamping the Transformer structure for handling extensive contexts, with corresponding future directions to push the frontier.
    \item In light of the extremely rapid growth of this field, 
    we build a repository that gathers relevant literature within this specific domain. We shall update it continuously to 
    keep pace with the latest advancements.
\end{itemize}

\noindent\textbf{Organization.} Section~\ref{sec: overview} gives an overview of long-context LLMs, including the preliminaries about objectives and stages for language modeling and critical components of Transformer-based LLMs, the structure limitation analyses for LLMs to deal with lengthy contexts and the taxonomy of existing efforts on advancing Transformer architecture. Then, we mainly delve into the discussion of each part of methodologies from the taxonomy in next five sections \ref{subsec: effcient_attention}$\sim$\ref{subsec: miscellaneous}, corresponding to related modules in Transformer architecture. In Section~\ref{sec: necessity_toolkit}, we also summarize the necessities for evaluating long-context capabilities and collect some popular optimization toolkits to augment LLMs' effectiveness and efficiency during training and inference. 
In Section~\ref{sec: discussion}, we explore the critical challenges and corresponding potential avenues lighted up by them, as well as draw insights from existing breakthroughs. Finally, Section~\ref{sec: conclusion} closes this survey with overarching conclusions regarding a panorama of the domain. 
    
    \section{Overview}\label{sec: overview}

\begin{figure*}[htbp]
    \centering
    \includegraphics[width=0.9\textwidth]{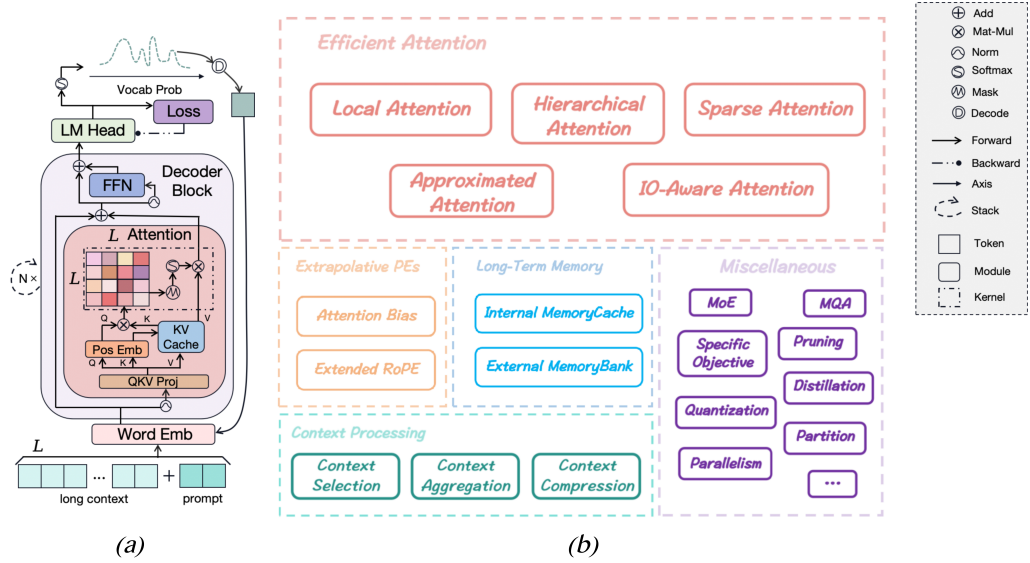}
    \captionof{figure}{The overview of 
    the survey: \figlabel{a} The typical architecture anatomy diagram of contemporary Transformer-based decoder-only LLMs, with the legend on the far top right; 
    \figlabel{b} The taxonomy of methodologies for enhancing Transformer architecture modules (corresponding to \figlabel{a} by color): \textcolor{attn_light_pink}{Efficient Attention} (submodule of \textcolor{attn_pink}{attention kernel}), \textcolor{memory_light_blue}{Long-Term Memory} (targeting \textcolor{memory_blue}{KV cache}), \textcolor{pe_light_orange}{Extrapolative PEs} (against the \textcolor{pe_orange}{positional embedding} module), \textcolor{context_light_green}{Context Processing} (related to \textcolor{context_green}{context pre/post-processing}), and \textcolor{mis_purple}{Miscellaneous} (general for the whole \textcolor{block_purple}{Decoder Block} as well as the \textcolor{loss_purple}{Loss} module).}
    \label{fig: overview}
\end{figure*}

In this section, we start with the preliminaries (Sec.~\ref{subsec: preliminaries}) for the fundamental language modeling objectives, typical modeling stages, as well as critical architecture modules in Transformer-based decoder-only LLMs, depicted in Fig.~\ref{fig: overview}(a). 
We then briefly analyze the architecture limitations when LLMs encounter extensive context windows (Sec.~\ref{subsec: limitation_analyses}). 
Finally, we present a 
taxonomy (Sec.~\ref{subsec: taxonomy}) of the different methods to enhance the long-context capabilities of LLMs through architectural innovations (cf.\ Fig.~\ref{fig: overview}(b)). 


\subsection{Preliminaries}\label{subsec: preliminaries}

\noindent\textbf{Language Modeling}. 
In a nutshell, (neural) language modeling aims to approximate the 
log-probability of the occurrence of any given text, denoted as $\log \mathrm{P}(X_{1: L}; \theta)$, where $\theta$ stands for the network parameters to be learned and $X_{1:L}$ 
comprises a sequence of length $L$ representing natural language including words, punctuation, mathematical symbols, etc. 
A significant practical hurdle for language modeling is 
\emph{curse of dimensionality}, i.e., 
the support of the probability distribution grows exponentially as $L$ increases. 
LLMs employ variations such as \emph{masked language modeling} (MLM) and \emph{causal language modeling} (CLM). The former is to predict masked tokens based on the bidirectional remaining unmasked tokens, i.e., 
\begin{equation}
       \mathrm{MLM}: 
       \arg\max\limits_{\theta}\; \sum\limits_{i \in \mathcal{M}} \log \mathrm{P}(x_i \mid X_{1: i-1, i+1:L};\theta) \label{eq: masked_language_modeling} 
\end{equation}
which maximizes the conditional probability of the $i$-th token $x_i$ given all the others, where $\mathcal{M}$ denotes the index set of the masked tokens. In contrast, the objective of CLM is to predict the next token, i.e., maximize the conditional probability of each token, given the unidirectional preceding ones 
\begin{equation}
        \mathrm{CLM}: \arg\max\limits_{\theta}\; \sum\limits_{i=1}^L \log \mathrm{P}(x_i \mid X_{1:i\sminus1};\theta) \label{eq: causal_language_modeling}
\end{equation}
%
In this setup, casual LLMs can effectively leverage the temporal dependencies inherent in natural language sequences, enabling LLMs to generate coherent and contextually relevant text.

\smallskip
\noindent\textbf{Modeling Stages}. Typically, 
LLMs often undergo a multi-stage modeling process. Initially, during the preprocessing stage, raw text data is segmented and tokenized into individual (sub)words, viz., \emph{tokens} predefined in a vocabulary, using algorithms, e.g., BPE~\cite{sennrich2015neural}. Then, in the pretraining stage, the model is trained on vast text corpora, with the MLM or CLM objectives, to capture semantic patterns and linguistic structures of natural language. Once pretrained, the model proceeds to the fine-tuning stage, where it is further trained with a few epochs on task-specific data with extra heads to learn sometimes. 
%
%
Finally, the finetuned model is deployed in downstream scenarios to predict expected answers in inference mode. Particularly, the casual LLMs are pretrained and finetuned with the same CLM objective but a different corpus. During the inference step, the model predicts from the probability distribution of the vocabulary by some decoding strategy such as greedy search, beam search, nucleus sampling~\cite{decoding-strategy}, to generate contextually coherent responses to prompts in a token-by-token autoregressive paradigm.


\smallskip 
\noindent \textbf{Decoder Block}. The vanilla Transformer architecture~\cite{vaswani2017attention} mainly comprises an Encoder and a Decoder, each stacked with multiple identical blocks. The skeleton of each block is mostly compatible with the one in Fig.~\ref{fig: overview}(a). In general, the first block takes the tokenized sequence encoded by a word embedding layer, followed by a \textit{multi-head scaled-dot self-attention} (MHA) layer with an attention mask corresponding to specific language modeling objectives and a feed-forward network (FFN) layer. Both the MHA and FFN layers are enriched with layer normalization~\cite{ba2016layer} and residual connections~\cite{he2016deep} at every entrance/exit of the block. Then, each higher-level block takes the output hidden states from the previous block as input, represents them with the MHA and FFN layers, and feeds them to the next block. The final hidden state outputted from the last block is fed into a linear layer called language modeling head, and the output logits is transformed into a probability distribution over the target vocabulary through the softmax operation. Note that the slight difference between the Encoder and Decoder blocks 
is that the latter additionally interfaces with the Encoder's output via a \emph{cross-attention} (CA) layer before feeding into the FFN layer. 

Such a binary structure was originally designed for sequence-to-sequence modeling in machine translation tasks. Subsequently, several variations have been proposed aiming at more general language modeling objectives such as MLM and CLM. The BERT series~\cite{devlin2018bert, liu2019roberta} harnesses only the Encoder with MLM to enhance bidirectional information, serving as a discriminative model. 
Conversely, the GPT series~\cite{radford2018improving, brown2020language, radford2019language} utilizes only the Decoder with CLM, focusing on unidirectional generative models. T5~\cite{raffel2020exploring} and BART~\cite{lewis2019bart} variants, however, treat each NLP task as a text-to-text conversion, leveraging both Encoder and Decoder. The decoder-only generative model architecture has recently become the predominant choice for current LLMs. Notable examples include 
%
Llama~\cite{touvron2023llama, touvron2023llama2}, 
OPT~\cite{zhang2022opt}, Bloom~\cite{workshop2022bloom},
GLM~\cite{du2021glm, zeng2022glm}, and
Mistral~\cite{jiang2023mistral, mixtral2023mistralai}, among others. 

\smallskip 
\noindent
\textbf{Attention Mechanism}. The attention mechanism~\cite{bahdanau2014neural}, as the core design of the Transformer implemented in the MHA layer, computes a weighted representation of each token in the input sequence based on its relevance to others. Specifically, as illustrated in Fig.~\ref{fig: overview}(a), the word-embedded token sequence $X \in \domain{L \times d_{in}}$, concatenating long contexts and user prompts with total length $L$, gives rises to three embedding matrices, i.e., a linear projection layer 
\textit{query} $Q \in \domain{L \times d_q}$, \textit{key} $K\in \domain{L\times d_k}$ and \textit{value} $V \in \domain{L\times d_v}$ 
\[ 
Q, K, V \defeq \mathrm{split}\left( X \times W_{q,k,v}\right), \quad W_{q,k,v} \in \domain{d_{in}\times (d_{q}+d_k+d_v)} \]
Then, for the attention kernel operations 
\[    
    P \defeq Q\times K^{\mathrm{T}}, \;\;
    A \defeq \mathrm{softmax}[\cfrac{P}{\sqrt{d_k}}\odot M], \;\;
    O \defeq (A\times V) \times W_o, \quad W_o \in \domain{d_v \times d_o} 
\]   
Namely, the unnormalized relevance matrix $P \in \domain{L \times L}$ 
each entry measures the relevance of the corresponding pair of tokens. 
The normalized attention score matrix $A \in \domain{L \times L}$ is computed as  a scaling operation by factor $\sqrt{d_k}$, an element-wise mask operation with $M\in \domain{L\times L}$, and a row-wise softmax. Finally, the output hidden states $O \in \domain{L\times d_o}$ are generated by a weighted sum of $V$ with attention weights in each row of $A$, usually with an extra linear transformation. 

Note that the embedding dimensions of $Q,K,V,O$ are not necessarily the same. 
Though subscripts are used to distinguish them for generality, by default we set $d=d_q=d_k=d_v=d_o$ in the rest of the paper. 
The mask matrix $M$ is typically used for masking padding tokens to align all batched input sequences and also applies casual mask operation of causal language modeling for generative LLMs. Furthermore, to capture diverse relationships, the model often employs multi-head attention instead of single-head one, performing the attention process in parallel with differently weighted $Q_h, K_h, V_h$ sets by dividing learnable parameters like $W_{q,k,v} \in \domain{d_{in} \times (3 \times d)}$ into $W_{q,k,v}^{mh} \in \domain{d_{in} \times (3 \times H \times d_{head})}$, where $H$ denotes the number of heads. Similar to embedding dimensions, the number of heads can be specific for $Q,K,V$, which vary in different LLMs, yet we consider them the same by default.


\smallskip 
\noindent
\textbf{Positional Embeddings}. 
Unlike recurrent neural networks (RNNs)~\cite{yu2019review}, Transformers process input tokens in parallel as a bag-of-words and lack an inherent sense of sequence order. To preserve the sequential information, the vanilla Transformer presents a novel Sinusoidal PE (SinPE)~\cite{vaswani2017attention}. 
\begin{equation}
    \mathrm{SinPE}(n) \defeq 
        \left[\begin{matrix}
            \sin(n\theta^0) \\
            \cos(n\theta^0) \\
            \sin(n\theta^1) \\
            \cos(n\theta^1) \\
            \vdots\\
            \sin(n\theta^{\frac{d}{2}\sminus1})\\
            \cos(n\theta^{\frac{d}{2}\sminus1})\\
        \end{matrix}\right], 
        \;\;where\;\; \theta = base^{-\frac{2}{d}}, \;n\in\{0,1,\cdots, L-1\} \label{eq: sinpe}
\end{equation}
%
Here $base$ is a large integer manually set as $10,000$ (according to the original paper without further explanation), and $d$ is the unit embedding dimension of hidden states. 

Some variants have recently emerged, including trainable embeddings~\cite{chen2021simple} to learn an embedding mapping and relative embeddings~\cite{shaw2018self} based on relative positions. For instance, Rotary PE (RoPE)~\cite{su2021roformer} applies a rotation operation on a complex field 
instead of an addition to $Q,K$ based on absolute positions, where it shares the same basis function as SinPE. 
\begin{equation}
    \mathrm{RoPE}(n) \defeq \left[\begin{matrix}
            R_n^{(0)}\\
            \; & R_n^{(1)}\\
            \; & \; & \ddots\\
            \; & \; & \; & R_n^{(\frac{d}{2}\sminus1)}\\
        \end{matrix}\right], \;\; where\;\; R_n^{(i)}\defeq\left[\begin{matrix}
            \cos(n\theta^i) & -\sin(n\theta^i)\\
            \sin(n\theta^i) & \cos(n\theta^i)\\
        \end{matrix}\right] \label{eq: rope} 
\end{equation}
Observe the properties 
\[
        ||R_i\bold q|| = ||\bold q||,\;\; P_{i,j} \defeq \langle R_i\bold q, R_j\bold k\rangle = \bold q^{\mathrm{T}}R_i^{\mathrm{T}}R_j\bold k = \bold q^{\mathrm{T}} R_{j- i} \bold k 
\]
RoPEs ensure the magnitude of $\bold q, \bold k$ remains unchanged (due to unitary transformation), and for every entry in $P$, i.e., each pair of $\bold q,\bold k$, will only be tagged with embeddings in terms of their relative distance in the sequence. RoPE provides a more stable scheme to handle longer sequences. It captures relative positional patterns with absolute position awareness, thus widely used in state-of-the-art open-source LLMs like LLama and GLM. 

It is worth noting that SinPEs are initially applied on the word embeddings before entering the Encoder or Decoder blocks by addition. In contrast, as shown in Fig.~\ref{fig: overview}(a), RoPEs are applied to $Q,K$ in each attention layer before the kernel operations by equivalent element-wise vector multiplication to save registered buffer memory.


\smallskip 
\noindent
\textbf{Key-Value Cache}. In a narrow sense, the Key-Value (KV) cache is a list of tensors that stores the $\bold k, \bold v$ embeddings for all previous tokens in the attention layer for each block, utilized and updated during the autoregressive generation process of causal LLMs. As shown in Fig.~\ref{fig: overview}(a),  before the first token is generated, all KV caches are initialized empty and will be filled with $L$ (key, value) pairs after the heave attention computation with $L$ queries and $L$ keys. Then, the first generated token will also be considered as input, extending the whole sequence to $L+1$ tokens. To avoid redundant calculations, the real input will contain only the latest generated token, deriving one new triplet of (query, key, value). But to compute equivalently, the new query has to attend and apply to all $L+1$ previous keys and values. Thus, the new (key, value) has to concatenate with past $L$ pairs stored in the KV cache and update themselves into it for the next generated token to attend. However, in a broad sense, we can consider the KV cache as the memory storage of LLMs, whose occupation grows linearly as the generated tokens increase. That directly causes one of the limitations below about the lack of efficient memory and suggests the approaches to enhance the \textit{long-term memory} mechanisms for LLMs in Section \ref{subsec: long-term_memory}.

\subsection{Limitations} \label{subsec: limitation_analyses}
 
\noindent
\textbf{Attention Complexity}. 
In typical scenarios where $L\gg d$, the computational complexity of MHA can be concisely summarized as follows. It involves $O(L^2d)$ time complexity, comprising $O(Ld^2)$ for QKV projection, $O(L^2d)$ for the computation of $P$, $O(L^2)$ for the \textit{softmax} operation to obtain $A$, $O(L^2d)$ for the multiplication of $A$ and $V$, and $O(Ld^2)$ for the output projection of $O$. It incurs $O(L^2)$ space complexity, involving $O(Ld)$ for embeddings of $Q, K, V, O$ and additional $O(L^2)$ buffers for storing weights $P$ and $A$. Consequently, both temporal and spatial computational costs exhibit a quadratic increase with the expansion of the sequence length, which can be burdensome for both training and inference.

\smallskip 
\noindent
\textbf{In-context Memory}. 
LLMs lack an explicit memory mechanism, relying solely on the KV cache to store representations of all previous tokens in a list. This design implies that once querying is completed in one call, the Transformer does not retain or recall any previous states or sequences in subsequent calls unless the entire history is reloaded token by token into the KV cache. Consequently, the Transformer possesses only an in-context working memory during each call, as opposed to an inherent memory mechanism such as Long Short-Term Memory (LSTM)~\cite{yu2019review}. This statelessness offers computational advantages in terms of parallelism but presents challenges in tasks like chatbot applications~\cite{kim2023chatgpt}, where long-term memory retention is essential.

\smallskip 
\noindent
\textbf{Max-Length Constraint}.
During the training phase, engineers typically need to determine a crucial hyperparameter \textit{max-length} ($L_{max}$ throughout this paper), which represents the upper bound on sequence length for any training sample in a batch. It is commonly set as 1K, 2K, or 4K based on the available computational resources to avoid Out-of-Memory~(OOM) errors on GPUs. 
However, during inference, LLMs service providers must either restrict the length of user prompts or automatically truncate them to align with the predefined $L_{max}$. 
Notice that none of the Transformer modules inherently require such restrictions since all learned weights depend solely on dimension sizes, hence Transformers theoretically can process sequences of any length. 
Unfortunately, current Language Models have shown noticeable performance degradation when handling input sequences exceeding $L_{max}$, often resulting in repetitive and implausible outputs.

\subsection{Taxonomy}\label{subsec: taxonomy}
There are multiple avenues to explore for advancing the Transformer structure to endow LLMs with long-context capabilities, such as reducing attention complexity during training, designing efficient memory mechanisms, and enhancing the ability for \textit{length extrapolation} where the model is trained on short sequences but tested on longer ones during inference~\cite{press2021train}. 
%
In this survey, we provide a comprehensive review of recent advancements in methodologies aimed at improving the long-context capabilities of LLMs throughout various stages. 
A taxonomy is given in Fig.~\ref{fig: overview}(b) where 
these methods are categorized into five main classes: 

\begin{itemize}
    \item \textbf{Efficient Attention} (Section~\ref{subsec: effcient_attention}). This class of methods focuses on implementing efficient attention mechanisms with reduced computational costs, even achieving linear-time complexity. Thereby $L_{max}$ in the pretraining stage can be increased, and so for the effective context length boundary of LLMs during inference.  
    \item \textbf{Long-Term Memory} (Section~\ref{subsec: long-term_memory}). This class of methods aims to design explicit memory mechanisms so the limitation of the in-context working memory can be addressed. 
    
    \item \textbf{Extrapolative PEs} (Section~\ref{subsec: extrapolative_pe}). This class of methods 
    improves the extrapolative properties of existing positional encoding schemes.

    \item \textbf{Context Processing} (Section~\ref{subsec: context_processing}). 
    This class of methods wraps off-the-shelf LLMs with additional context pre/postprocessing. They ensure that the input fed to LLMs in each call always meets the maximum length requirement and breaks the context window limit by introducing multiple calling overheads.
    
    \item \textbf{Miscellaneous} (Section~\ref{subsec: miscellaneous}). This class 
    includes various 
    methods that do not naturally fit into the previous four categories, offering a broader perspective on advancing long-context capabilities in LLMs.
\end{itemize}
    

\section{Efficient Attention} \label{subsec: effcient_attention}
The first category of methods is to optimize attention mechanisms, especially the kernel operations that are the computational bottleneck of the Transformer. 
This approach enables the expansion of the context length boundary for LLMs during inference by directly increasing the hyperparameter $L_{max}$ in the pretraining stage. We further categorize these methods into five distinct strategies, each with a specific focus: Local Attention (Sec.~\ref{subsubsec: local_attn}), Hierarchical Attention  (Sec.~\ref{subsubsec: hierarchical_attn}), Sparse Attention (Sec.~\ref{subsubsec: sparse_attn}), Approximated Attention (Sec.~\ref{subsubsec: approx_attn}) and IO-Aware Attention (Sec.~\ref{subsubsec: ioaware_attn}).

\subsection{Local Attention}\label{subsubsec: local_attn}
\begin{figure*}[htbp]
    \centering
    \includegraphics[width=0.7\textwidth]{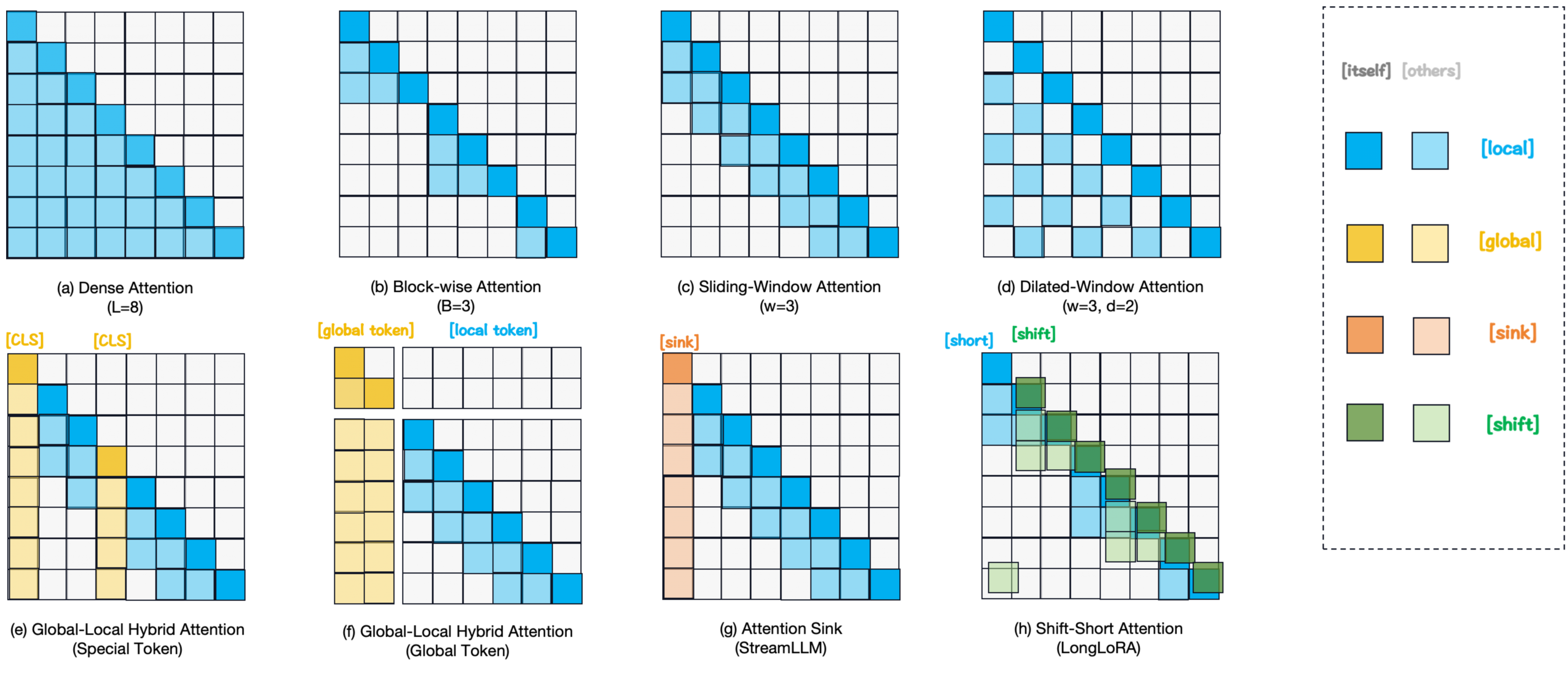}
    \captionof{figure}{The visualization of various typical local causal attention mechanisms. As the legend on the right indicates, tokens are distinguished by colors, with shades denoting attention to themselves (darker) or attention to the preceding others (lighter).} 
    \label{fig: local_attention}
\end{figure*}

The traditional attention mechanism is characterized by its global and full attention nature, wherein every token is expected to attend to every other token, resulting in quadratic time and space complexities. Considering the significance of local context in certain applications~\cite{yang2021context}, various approaches have been introduced to implement local attention mechanisms in recent years. These mechanisms restrict each token's attention to its neighboring tokens only. Variations of these approaches arise from the heuristics 
to determine what qualifies as a token's neighbor, as depicted in Fig.~\ref{fig: local_attention}.

\noindent\textbf{Block-wise Attention}.
One straightforward approach to local attention involves segmenting the input sequence into non-overlapping blocks. As proposed in BlockBERT~\cite{qiu2019blockwise}, tokens are restricted to attending only to others within the same block of fixed size $B$. This block-wise attention requires full attention calculations within each $B\times B$ block for $\frac{L}{B}$ iterations, resulting in time complexity $O(LBd)$ and memory complexity $O(LB)$. However, this approach limits the global receptive field, potentially hindering long-term dependency modeling. To mitigate this, Bi-BloSAN~\cite{shen2018bi} introduces inter-block attention for capturing long-range dependencies. Sinkhorn~\cite{tay2020sparse} employs a differentiable ranking network to sort blocks, enabling quasi-global receptive fields. SPADE~\cite{zuo2022efficient} augments state space models (SSMs)~\cite{gu2021efficiently} to address long-range dependency limitations. Additionally, Landmark Attention~\cite{mohtashami2023landmark} introduces a \textit{landmark token} for each block to select relevant neighbors, facilitating block-wise representations. In the fine-tuning stage, LongLoRA~\cite{chen2023longlora} introduces shift short attention (S$^2$-Attn) atop LoRA~\cite{hu2021lora}, shifting tokens by half the block size in half of the attention heads to ensure information flow between neighboring blocks.

\noindent\textbf{Sliding Window Attention}. 
Inspired by convolutional neural networks (CNNs)~\cite{lecun1998gradient, krizhevsky2012imagenet}, another approach is to use sliding-window techniques, as demonstrated in Longformer~\cite{beltagy2020longformer}. In this method, each token is assigned a consecutive fixed window and is allowed to attend only to the previous adjacent $w \ll L$ tokens as its neighbors. To extend the receptive field similar to dilated convolution~\cite{yu2015multi}, the window is dilated with gaps of size dilation $d$, enabling each token to attend to tokens as far as $w\times d+1$ away. To aggregate global information without additional computation, global attention is also applied to a few pre-selected positions where special tokens like \texttt{[CLS]} are located, decreasing computation complexity to $O(Lwd)$. Besides, Funnel-Transformer~\cite{dai2020funnel} employs a strided mean pooling strategy on each window to compress the sequence dimension of hidden layers, while Sequence-AltUp~\cite{baykal2023alternating} further captures contextual information in skipped tokens with its lightweight predictor.


\noindent\textbf{Global-Local Hybrid Attention}.
A similar global-local attention mechanism is adopted in ETC~\cite{ainslie2020etc} and LongT5~\cite{guo2021longt5}, which construct auxiliary global tokens explicitly or implicitly to represent segment information with global attention, while applying local attention only to source tokens (This hierarchical organization of attention receptive fields is further discussed in Sec.~\ref{subsubsec: hierarchical_attn}). To avoid tuning, LongLM~\cite{jin2024llm} employs grouped attention for out-of-window tokens and standard attention for those within the
neighbor window. StreamLLM~\cite{xiao2023efficient} observes the phenomenon of \textit{attention sink} that maintaining KV of initial tokens during inference largely recovers sliding window attention performance, and adding a placeholder token during pretraining further improves streaming deployment. This phenomenon arises from the strong attention towards initial tokens as a "sink" even if they are not semantically important. Similarly, Lm-infinite~\cite{han2023lm} proposes a $\Lambda\sminus$shaped mask and positional distance constraint to keep attending to starting tokens.

\noindent\textbf{LSH Attention}. 
Besides the direct positional adjacency, Reformer~\cite{kitaev2020reformer} utilizes a neighbor token selection mechanism based on k-Nearest-Neighbor (kNN) and Locality-Sensitive Hashing (LSH) algorithms~\cite{jafari2021survey}. LSH attention allows each query $\bold q_i$ to attend to a set of keys $S_i \defeq \{\bold k_{j\le i}: h(\bold q_i) = h(\bold k_j)\}$ within a single hash bucket. The hashing function $h$ is designed to assign the same hash with high probability to two vectors that are similar and vice versa. This approach ensures that each token can access a fixed number $K \ll L$ of neighboring keys, and the primary computational cost of LSH attention arises from bucket sorting, with a complexity of $O(L\log Ld)$.

\subsection{Hierarchical Attention}\label{subsubsec: hierarchical_attn}
The global token techniques~\cite{beltagy2020longformer, ainslie2020etc, guo2021longt5} and the inter-block attention~\cite{shen2018bi} mentioned above are essentially introducing hierarchical features to self-attention to compensate with more global information from the higher-level attention while keeping the low computation cost from the low-level local attention at the same time. From this view, more work has explored various hierarchical mechanisms that introduce a structured hierarchy into self-attention, leveraging higher-level global information and lower-level local attention for multi-scaled contextual receptive fields.

\noindent\textbf{Two-Level Hierarchy}. 
HAN~\cite{yang2016hierarchical} pioneers the use of a two-level attention mechanism. It first applies self-attention to word features to obtain a sentence representation, then employs self-attention on sentence-level features to generate document-level features. This hierarchical approach improves efficiency and performance in document classification tasks. Subsequently, similar hierarchical attention mechanisms have led to significant advancements in other document-level tasks, including machine translation~\cite{miculicich2018document, rohde2021hierarchical, wu2020lite}, and document summarization~\cite{zhang2019hibert, zhang2022hegel, cohan2018discourse}.

\noindent\textbf{Multi-Level Hierarchy}. 
In contrast to the typical binary level structure above, BPT~\cite{ye2019bp} introduces a more elaborated fine-to-coarse attention mechanism that operates on multi-scale spans via binary partitioning. Token nodes can attend to smaller-scale spans for close context and to larger-scale spans for   distant context. This approach formalizes the hierarchical structure as a graph neural network and updates it using graph self-attention~\cite{velickovic2017graph}. A simpler variation is adopted in Adaptive Span Transformer~\cite{sukhbaatar2019adaptive}, which employs a soft attention masking function to non-increasingly map relative distances to real values in the range $[0, 1]$. This function controls the span of attention for each head, allowing the model to attend to different context spans. 

Building on the prior studies~\cite{khandelwal2018sharp, abreu2019hierarchical} that indicate a hierarchical low-rank structure in attention matrix across NLP tasks,
H-Transformer-1D~\cite{zhu2021h} introduces hierarchical attention, partitioning the matrix into blocks with varied low-rank ranges for diverse approximation levels. This reduces runtime and memory complexity to $O(Ld)$, where the number of hierarchy levels $M$ is typically set to $\log{(L/2)}$. Viewing full-attention as a conditional expectation over embeddings at each location, Combiner~\cite{ren2021combiner} approximates this conditional distribution with structured factorization on token regions. Tokens can then attend to others either directly or through indirect attention to \textit{abstractions}, which are conditional expectations from corresponding factorized local regions. This approach also leverages sparse attention patterns, as will be discussed in Sec.~\ref{subsubsec: sparse_attn}, to provide sub-quadratic low computation and memory complexity while maintaining full-attention expressiveness.

Generally speaking, hierarchical attention mechanisms derive from the same principles of contextual locality present in natural languages as local attention. However, they incorporate a more elaborated structure, often designed heuristically, to strike a balance between capturing long-range contextual dependencies and maintaining low-level computational efficiency.

\subsection{Sparse Attention}\label{subsubsec: sparse_attn}

\begin{figure*}[htbp]
    \centering
    \includegraphics[width=0.8\linewidth]{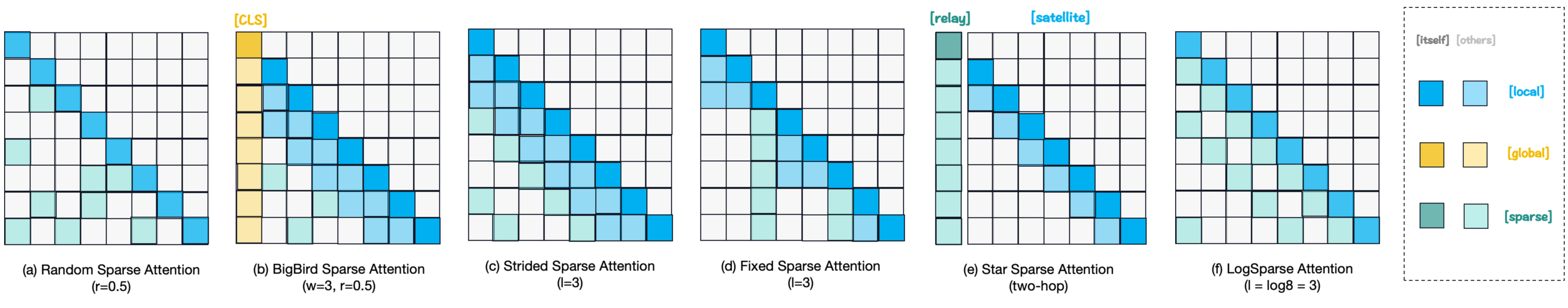}
    \captionof{figure}{The visualization of some typical causal sparse attention patterns. The legend on the right distinguishes token types based on their colors, where darker shades indicate attending to themselves while lighter ones represent attention to other previous tokens.} 
    \label{fig: sparse_attention}
\end{figure*}

While some approaches have introduced heuristics for achieving locality and hierarchical structure within self-attention, another direction explores the sparsity patterns inherent in full attention matrices~\cite{child2019generating, yun2020n}. These methods aim to introduce a sparse attention mask, denoted as $M_{\mathcal{S}}$, where each row $i$ assigns a sparse set of indices $\mathcal{S}_i \subseteq \{j | j < i\}$ that the $i$-th token attends to. These sparsity-based attention mechanisms offer both computational efficiency and the ability to capture global context information~\cite{microsoft2020deepspeed}. Fig.~\ref{fig: sparse_attention} provides a visualization of these sparse attention mechanisms.

\textbf{Fixed Sparsity Patterns}.
To start with Sparse Transformer~\cite{child2019generating}, it draws inspiration from attention patterns learned on CIFAR-10~\cite{krizhevsky2009learning} and proposes a row-column factorized attention scheme. This approach results in faster computations while still maintaining global context awareness. Formally, it employs a chosen stride $l$ that is close to $\sqrt{L}$. Each query $\bold q_i$ applies one \textit{row attention} for local context information (i.e., local attention) and another \textit{column attention} that summarizes previous locations and propagates information to all future tokens, resembling a form of global attention. The authors provide two specific patterns for row and column attention, corresponding to the \textit{stride} and \textit{fixed} ones respectively illustrated in Fig.~\ref{fig: sparse_attention}(c) and Fig.~\ref{fig: sparse_attention}(d). This strategy reduces the total computational complexity to $O(L\sqrt Ld)$. The intuition of Sparse Transformer is to attribute a stride $l\approx \sqrt L$ to equally distribute $\sqrt L$ tokens to attend to. In contrast, LogSparse~\cite{li2019enhancing} employs an exponentially sparse attention pattern by dispatching only $\log L$ tokens for each location to attend to. This method ensures that any pair of tokens can eventually exchange information with each other through a path spanning $\log L$ layers. This results in an overall memory usage of $O(L(\log L)^2)$. The recent LongNet~\cite{ding2023longnet} further improves computational efficiency by introducing \textit{dilated attention}, which expands the attentive field exponentially as the distance between tokens increases. It incorporates mixed dilate rates to model both long and short-range dependencies, ultimately reducing the computation complexity to $O(Ld)$ while successfully scaling to sequences of up to one billion tokens.


\textbf{Adaptive Sparsity Patterns}.
Instead of fixed sparse indices set only dependent on locations, some approaches seek sparsity adaptively in a learnable manner, taking into account embedding values. Expire-Span~\cite{sukhbaatar2021not} introduces a learnable scalar in the range $[0, 1]$ for each previous token, allowing the model to retain tokens with the most important information while expiring those that are no longer relevant, similar to the \textit{forget gate} in LSTM-based RNNs~\cite{yu2019review}. Routing Transformer~\cite{roy2021efficient} leverages \textit{k}-means clustering to identify the top-\textit{k} most relevant \textit{centroid} vectors in $Q, K$ and assigns each query to the keys with the same cluster membership, reducing the overall complexity of attention to $O(L\sqrt Ld)$. Inspired by Differentiable
Architecture Search (DARTS)~\cite{liu2018darts}, SparseBERT~\cite{shi2021sparsebert} introduces a differentiable attention mask using Gumbel relaxation techniques~\cite{maddison2016concrete}, allowing the model to learn to guide attention pattern selection by importance. It incorporates a predefined sparsity ratio $\rho$, resulting in computational complexity of $O((1-\rho^2)L^2d)$.

\textbf{Graph Sparsification}.
Furthermore, some other works treat full attention as a fully connected graph, with nodes representing embeddings of each token and edges denoting connections through attention. These approaches frame sparsity as a \textit{graph sparsification problem}. For instance, Star-Transformer~\cite{guo2019star} introduces a star-shaped topology, where each \textit{satellite} node attends to local neighbors with a ring connection and a virtual \textit{relay} node with the radial connection. In contrast, BigBird~\cite{zaheer2020big} incorporates sparsity based on random graph theory, allowing each query to attend to a random number of keys with a fixed probability. It also absorbs sliding-window local attention with window size $w$ and global token techniques in its design. This approach reduces the quadratic dependency to linear complexity, specifically $O((w+r+1)Ld)$, stacked with three efficient attention mechanisms.

\subsection{Approximated Attention}\label{subsubsec: approx_attn}
In addition to the heuristics to restrict full attention computation, some research 
approximates attention
based on the sparsity or low-rankness of attention matrices 
with linear complexity, albeit at the cost of precision. We introduce several of these approximation techniques below.

Notice that we do not distinguish whether the methods are employed for BERT-like encoder-only LLMs or GPT-like decoder-only LLMs in previous sections since most of them can be trivially transferred from the BERT setting to the GPT setting with a causal attention mask. However, the casual mask is often non-trivial for many approximation strategies. So to facilitate later discussions, we first define a general weighted causal function $\xi_{\bold w}$ in Eq.~\ref{eq: causal_function}, where $\bold w \in \domain{L}$ represents a weights vector for each row. This function will substitute the causal attention mask operation. Thus, we omit the mask $M$ in all attention equations below for simplification.

 \begin{equation}
        \xi_{\bold w}(Q,K,V) \defeq \left[w_i\cdot \bold  q_i\transpose \sum\limits_{j=1}^i \bold k_j \bold{v}_j\transpose \right]_{i=1}^L\label{eq: causal_function}
 \end{equation}

\noindent\textbf{Low-Rank Approximation}.  Linformer~\cite{wang2020linformer} employs Singular Value Decomposition (SVD) to approximate the attention matrix $A$ with a low-rank matrix $\widetilde{A}$
This approach involves two learnable projection matrices $E$ and $F$ of dimensions $L \times k$, where $k = O(\frac{d}{\epsilon^2}) \ll L$. The process includes projecting $K, V$ using $E, F$ respectively, followed by standard MHA kernel on $Q$ with the projected $\widetilde{K}, \widetilde{V}$. According to the properties proved in Linformer, this low-rank technique approximates full attention with linear complexity $O(Lkd)$ while allowing for an error of $\epsilon$. 


\noindent \textbf{Nested Attention}. 
Luna~\cite{ma2021luna} decouples the attention kernel into two nested attention approaches, both of which have linear complexity relative to $L$. Specifically, it firstly applies \textit{pack attention} as Eq.~\ref{eq: pack_attention} to get \textit{packed context} $\widetilde S$, where $S$ is an extra side-input sequence with constant length $k\ll L$. The activation function $\mathrm{elu}(\cdot)$ is the exponential linear unit~\cite{clevert2015fast}. Then it secondly applies \textit{unpack attention} as Eq.~\ref{eq: unpack_attention} to get \textit{unpacked output} $\widetilde O$, with the causal mask function defined in Eq.~\ref{eq: causal_function}. Afterward, they pass $\widetilde O$ and $\widetilde S$ to the next attention layer, denoted as $X$ and $S$, to propagate packed contextual information via $S$ without leakage of future information. Notice that the pack attention can be regarded as a generalization of the linear projection in Linformer~\cite{wang2020linformer} with the same complexity $O(Lkd)$ but the advantage to model sequences with various lengths since the projection matrices are not dependent to $L$ as projection matrices $E$ and $F$.

    \begin{align}
        &A_{s} \defeq \mathrm{elu}\left( \frac{Q_s \times K\transpose}{\sqrt{d_k}} \right), \quad\widetilde S \defeq A_{s}\times V, \;\;where\;\; Q_s \defeq S\times W_q\label{eq: pack_attention}\\
        &A_{u} \defeq \mathrm{softmax}\left( \xi_{\bold w_{inv}}(Q, V, A_{s}\transpose) \right),\;
        \widetilde O \defeq \xi_{\bold w_{inv}}(A_{u}, A_{s}\transpose, V) \label{eq: unpack_attention},\;\;where\;\; \bold w_{inv} \defeq \left[ i^{\sminus1} \right]_{i=1}^L\
    \end{align}

\noindent\textbf{Kernelized Approximation}.
Except for low-rankness prior, some works are based on generalized kernelizable attention in Eq.~\ref{eq: factorization_attention}, where the kernel function $\mathcal{K}(\cdot,\cdot): \domain{d}\times \domain{d}\rightarrow R_+$ is applied row-wise to each pair of $\bold q_i, \bold k_j$ in $Q,K$, and $D$ is the normalization factor defined in Eq.~\ref{eq: factorization_normalization}. From this view, the vanilla softmax attention implements a specific kernel function as $\mathcal{K}(Q,K) = \exp(\frac{QK\transpose}{\sqrt{d_k}})$, which explicitly derives a $L\times L$ attention matrix. But suppose we carefully choose another kernel function to be factorizable as the condition in the second step of Eq.~\ref{eq: factorization_attention} and \ref{eq: factorization_normalization}, then simply apply the associative property. In that case, we can compute matrix multiplication of $K,V$ and $K, \bold 1_L$ ahead with lower complexity $O(Ld^2)$. However, the drawback is that similar to Luna~\cite{ma2021luna}, one has to compute the matrix multiplication iteratively across each query $\bold q_i$, which does not fully make use of the parallelism, as shown in the third step of Eq.~\ref{eq: factorization_attention} and Eq.~\ref{eq: factorization_normalization}.
%
    \begin{align}
        & O \defeq \left(D^{-1}\times\mathcal{K}(Q, K)\right)\times V
        \xlongequal[associative]{\mathcal{K}(Q, K)=\widetilde Q\times \widetilde K^\mathrm{T}} D^{-1}\times\widetilde Q \times \left(\widetilde K^\mathrm{T} \times V\right) 
        \xlongequal[]{causal} D^{-1}\times \xi_{\bold 1_L}(\widetilde Q, \widetilde K, V)\label{eq: factorization_attention}\\
        & where\;\;\; D \defeq\mathrm{diag}\left[\mathcal{K}(Q, K) \times \bold 1_L\right]
        \xlongequal[associative]{\mathcal{K}(Q, K)=\widetilde Q\times \widetilde K^\mathrm{T}} \mathrm{diag}\left[\widetilde Q \times\left(\widetilde K\times \bold 1_L\right)\right]
        \xlongequal[]{causal} \mathrm{diag}\left[ \xi_{\bold 1_L}(\widetilde Q, \widetilde K, \bold 1_L)\right]\label{eq: factorization_normalization}
    \end{align}

For instance, Linear Transformer~\cite{katharopoulos2020transformers} designs a simple feature map $\varphi_{Li}$ based on the $elu$ kernel in Eq.~\ref{eq: elu_kernel}. It avoids the quadratic attention matrix and reduces the time (resp.\ space) complexity to $O(Ld^2)$ (resp.\ $O(Ld)$). 
In contrast, Performer~\cite{choromanski2020rethinking} achieves unbiased and low-variance estimation based on orthogonal random features (ORFs) mapping $\varphi_{Pe}(\cdot):\domain{d} \rightarrow \domain{r}$ in Eq.~\ref{eq: orf_kernel}, where $r = m \times l \ll L$,  $b_1,\cdots,b_l: \domain{} \rightarrow \domain{}$ are $l$ basis functions, $h: \domain{d}\rightarrow\domain{}_+$ is certain magnitude measurement, and $\omega_1,\cdots,\omega_m \in \domain{d}$ are $m$ orthogonal random features $i.i.d$ sampled from a distribution $\mathcal{D} \in \mathcal{P}(\domain{d})$. 
Performer provides multiple configurations of these parameterized functions, like $h\defeq 1, l\defeq1, \mathcal{D}\defeq \mathcal{N}(0,\mathbf{I}_d)$ as PNG-kernels~\cite{choromanski2017unreasonable}, and $h\defeq 1, l\defeq2, b_1\defeq \sin, b_2\defeq cos,\mathcal{D}\defeq \mathcal{N}(0,\sigma^2\mathbf{I}_d)$ 
as shift-invariant Gaussian kernel~\cite{rahimi2007random}, adopted in RFA~\cite{peng2021random}. Hence, the time (resp.\ space) complexity can be reduced to $O(Lrd)$ (resp. $O((d+r)L)$).
%
    \begin{align}
        &\mathcal{K}_{Li}(\bold q,\bold k) \defeq \varphi_{Li}(\bold q)\times \varphi_{Li}(\bold k)^\mathrm{T}, \;\;\text{where}\;\; \varphi_{Li}(\bold x) = \mathrm{elu}(\bold x) + 1  \label{eq: elu_kernel}\\
        &\mathcal{K}_{Pe}(\bold q,\bold k) \defeq \mathbb{E}_{\omega}\left[\varphi_{Pe}(\bold q)\times \varphi_{Pe}(\bold k)^\mathrm{T}\right], \;\text{where}\; \varphi_{Pe}(\bold x) = \frac{h(\bold x)}{\sqrt{m}} \left[b_1(\omega_1\transpose\bold x),\ldots,b_1(\omega_m\transpose\bold x),\ldots, b_l(\omega_1\transpose\bold x),\ldots,b_l(\omega_m\transpose\bold x) \right] \label{eq: orf_kernel}
    \end{align}

The works above pioneer a novel approach to enhancing attention efficiency by treating it as a kernel machine, spawning further kernel strategies such as Fourier Attention~\cite{nguyen2022fourierformer} and Primal Attention~\cite{chen2023primalattention}.

\noindent\textbf{Sparse-Kernelized Hybrid}.
Furthermore, inspired by Robust-PCA~\cite{candes2011robust}, the recent Scatterbrain~\cite{chen2021scatterbrain} provides a more accurate yet efficient approximation by combining LSH-based sparse matrices $S$ like Reformer's~\cite{kitaev2020reformer} and low-rank kernelized decomposition with randomized feature maps like Performer's~\cite{choromanski2020rethinking}, as simplified in Eq.~\ref{eq: scatterbrain}, where we omitted the normalization step and the causal mask applying the function. 
%
    \begin{align}
        &\widetilde O \defeq \left( \widetilde Q \times \widetilde K\transpose + S \right)\times V = \widetilde Q \times (\widetilde K\transpose\times V) + S\times V  \label{eq: scatterbrain}
    \end{align}
%
Not only does this method unify two approximation techniques to achieve linear time complexity of $O(Lrd)$ with higher precision, but it also offers flexibility to leverage various low-rank and sparse approximation methods as sub-components, more than just the example combination of Reformer and Performer.

\noindent\subsection{IO-Aware Attention}\label{subsubsec: ioaware_attn}
All of the methods above in pursuit of efficient attention can be considered as trading off high attention quality for low computation complexity, based on some theoretical or empirical properties of attention matrix and NLP tasks, including locality, sparsity, low-rankness, and other heuristic or mathematical tricks. In comparison, these IO-aware attention mechanisms below collectively represent efforts to optimize attention computations by considering the memory bottleneck while preserving the exactness of attention kernel calculations.

\noindent\textbf{Memory-Efficient Attention}.
This simple method is firstly proposed in~\cite{rabe2021self}, which utilizes the \emph{lazy softmax} algorithm~\cite{jang2019mnnfast} and tracks normalization factor to compute standard and numerically stable attention by sequentially processing each single/chunked-query attention. Such a simple method only needs constant working memory with respect to sequence length, while the time complexity is still quadratic. 

\noindent\textbf{Flash Attention}.
Furthermore,
the recent work Flash Attention~\cite{dao2022flashattention, dao2023flashattention} manages to reduce time and memory consumption while applying exact attention computation by making it IO-aware between GPU high bandwidth memory~(HBM) and GPU on-chip SRAM. It has already reached wide adoption and has been incorporated directly into Pytorch v2.0~\cite{pytorch2.0}. To be more specific, for forward pass, it utilizes the tiling technique~\cite{xu2009tiling} to decompose large softmax attention into smaller blocks, loading block by block from HBM~\cite{jia2018dissecting} to SRAM, to perform all the attention computation steps on-chip to reduce HBM accesses, and incrementally update the output back to HBM by rescaling, leveraging the online softmax technique~\cite{milakov2018online}. As Eq.~\ref{eq: tiling} illustrates, $P^{(n)} \defeq Q_{i_n}\times K_{j_n}\transpose \in \domain{B_r \times B_c}$ denotes one pair of block-wise matrix multiplication from $Q,K$, which is tiny enough to be loaded and computed in SRAM, and the rescaling factor $\alpha^{(n)}$ only comprises some statistics about $P^{(n)}$. As for backward pass in training stages, it can both decrease the required memory and speed up due to reduced HBM accesses as well, by just storing the output $O$ and normalization statistics to recompute the intermediate results $P,A$ in SRAM, similar to the gradient checkpointing technique~\cite{chen2016training, sohoni2019low}. The authors analyze that flash attention only requires $O(L^2d^2M^{-1})$ HBM accesses compared to standard $O(Ld+L^2)$, where $M \gg d^2$ is the size of SRAM, which leads to both faster execution (up to  7.6× speedup on GPT2~\cite{radford2019language}) and lower memory footprint (up to 20× more memory efficient), according to the experiments results~\cite{dao2022flashattention}.
%
%
    \begin{align}
        &O \defeq \mathrm{softmax}\left( \left[\begin{matrix} P^{(1)} & P^{(2)} \end{matrix} \right]  \right) \left[\begin{matrix} V^{(1)} \\ V^{(2)} \end{matrix} \right]
        = \alpha^{(1)} \mathrm{softmax}(P^{(1)}) V^{(1)} + \alpha^{(2)} \mathrm{softmax}(P^{(2)}) V^{(2)}\label{eq: tiling}
    \end{align}

\noindent\textbf{SCFA}. Although Flash Attention can be easily extended to support block-sparse structures~\cite{dao2022flashattention}, it may lack flexibility for handling other sparse strategies with irregular structures and arbitrary attention masks. The effort of SCFA~\cite{pagliardini2023faster} extends the Flash Attention GPU kernel to accommodate a broad range of attention sparsity patterns, including key/query dropping and hashing-based attention like Reformer~\cite{kitaev2020reformer}. This extension leads to a training speedup of 2.0 to 3.3 times without sacrificing perplexity, according to the report from the paper.

\noindent\textbf{Paged Attention}.
While Flash Attention has effectively tackled the training memory bottleneck, LLMs still face challenges related to the memory consumption of the KV cache during inference, which grows dynamically with batched requests. Recognizing the memory wastage due to fragmentation and redundancy, vLLM proposes Paged Attention~\cite{kwon2023efficient}. This technique efficiently manages KV cache memory to minimize waste and allows flexible sharing across batched requests, drawing inspiration from memory paging techniques~\cite{kilburn1962one} in virtual memory operating systems~\cite{denning1970virtual}.

\section{Long-Term Memory}\label{subsec: long-term_memory}
The Transformer architecture often struggles with capturing long-term dependencies due to in-context working memory, as highlighted in Sec.~\ref{subsec: limitation_analyses}. Researchers have explored two main avenues to address this challenge without compromising the advantages of full attention. Inspired by RNNs, some introduced recurrent mechanisms into attention by incorporating internal memory caches accessible through attention layers. This approach enables the model to maintain and retrieve information over longer sequences, compensating for the inherent lack of built-in long-term memory. An alternative approach involves leveraging existing models as interfaces to external knowledge bases, such as specific documents or datasets. During inference, the model can read from these knowledge bases to enrich its contextual input and write to them from the user's response to refresh its long-term memory. By integrating external knowledge in this manner, the model gains access to a broader range of context, enhancing its ability to handle long-term dependencies effectively.

\subsection{Internal MemoryCache}

Recalling the temporality of natural language representations instead of the success of full parallelism in Transformer, we introduce the concept of \textit{Internal MemoryCache} based on recurrence mechanisms. It divides the long text into a stream of fixed-length segments and enhances the query $Q^{n}_{t}$ of the current $t$-th segment in the $n$-th layer with more contextual information $\widetilde K^{n}_{t},\widetilde V^{n}_{t}$. This contextual information is obtained from cached or distilled information from previous segments, stored in a memory cache denoted as $Mem$, as shown in Eq.~\ref{eq: recurrence}. To facilitate later explanations, we assume that each segment has the same length $l$, and the models consist of $N$ layers of transformer blocks. The operator~$[\circ ]$ represents the concatenation operation along the length dimension. It is worth noting that the variables in the memory cache $Mem$ are usually detached from the computation graph, eliminating the need for gradient computation, which we denote with a hat accent, such as $\widehat X$.
%
    \begin{align}
        & Q^{n}_{t}, \;\widetilde K^{n}_{t}, \;\widetilde V^{n}_{t} \defeq X^{n}_{t}W_q, \;\widetilde X^{n}_{t}W_k, \;\widetilde X^{n}_{t}W_v,
        \;\;where\;\; X^{n}_{t} \defeq O^{n-1}_{t}, \; \widetilde X^{n}_{t} \defeq \left[ \mathrm{Mem}(n,t, \ldots) \circ O^{n-1}_{t}  \right]\label{eq: recurrence}
    \end{align}

\noindent\textbf{Segment-Level Recurrence}.
The segment-level recurrence is initially introduced into Transformer from Transformer-XL~\cite{dai2019transformer}. As illustrated in Eq.~\ref{eq: transformer_xl}, it caches the output of $m$ previous consecutive segments in the last layer and concatenates them into the current segment in the present layer to extend the context for the current query. Such a mechanism allows for extending the largest possible dependency distance to $O(Nml)$, where $m$ can be set as far as GPU memory allows. Building upon Transformer-XL, Segatron~\cite{bai2021segatron} introduces the segment-aware mechanism by enhancing the token-level PEs combined with sentence-level and even paragraph-level ones. To further extend the dependency with multi-grained memory caching, Compressive Transformer~\cite{rae2019compressive} stores the first FIFO fine-grained memory queue for $m_1$ previous segments as Transformer-XL does. However, instead of discarding old memory, it applies a \textit{compression function} $f_c$ with the rate $c$ to compress it along the length dimension and pushes it into a secondary FIFO coarse-grained \textit{compressive memory} queue of size $m_2$. Combining these two types of memories, one can obtain the longest context dependency as $O(Nl(m_1 + c m_2))$, as shown in Eq.~\ref{eq: compressive}.

    \begin{align}
        & \mathrm{Mem}_{XL}(n,t,m) \defeq \left[\widehat O^{n-1}_{t-m} \circ \ldots \circ \widehat O^{n-1}_{t-1}\right]\label{eq: transformer_xl}\\
        & \mathrm{Mem}_{Comp}(n,t,m_1,m_2,c) \defeq \left[\mathrm{Mem}_{f_c} \circ  \mathrm{Mem}_{XL}(n,t,m_1)\right], \label{eq: compressive}\\
        &where\;\; \mathrm{Mem}_{f_c} \defeq \left[ f_c(\widehat O^{n-1}_{t-m_1-m_2}) \circ\ldots \circ  f_c(\widehat O^{n-1}_ {t-m_1-1})\right]\notag
    \end{align}

\noindent\textbf{Retrospective Recurrence}.
Notice that both Transformer-XL and Compressive Transformer deploy a shifting-one-layer-downwards recurrence by default, thus the maximum effective context length is limited by $N$. To address it, similar to Feedback Transformer~\cite{fan2020addressing}, ERNIE-Doc~\cite{ding2020ernie} proposes an enhanced recurrence mechanism, a drop-in replacement by concatenating the output hidden states of previous segments in the same layer, instead of the last layer, simply formalized as Eq.~\ref{eq: ernie_doc}. In this manner, not only the maximum effective context length can be implicitly expanded, but also the past higher-level representations can be exploited to enrich future lower-level representations as well. Additionally, it employs a \textit{retrospective feed} mechanism by feeding the segments twice, where the first time only skims each segment while the second one retrospects to enable bi-directional information flow, which resembles the mechanism in READTWICE~\cite{zemlyanskiy2021readtwice}.

 \begin{align}
    & \mathrm{Mem}_{\infty}\defeq \widetilde{X}(s) = B\transpose \Phi(s), \;\;s.t.\;\; \widetilde X(s_i) \approx X_i, \;s_i \defeq i / L, \;\forall i \in [1,\ldots ,L]\label{eq: infty}\\
    & \mathrm{Mem}_{Ernie}(n,t) \defeq \widehat O^{n}_{t-1}\label{eq: ernie_doc}
\end{align}

\noindent\textbf{Continuous-Signal Memory}.
To draw a comparison with LSTM~\cite{yu2019review}, we can view the compressed memory in Compressive Transformer as a finite-sized discrete version of the long-term cell memory in LSTM, while the first queue stores the short-term one. To achieve unbounded long-term memory like LSTM,  as Eq.~\ref{eq: infty} suggests, $\infty$-former~\cite{martins2021infty} transfers the $L$ token-wise discrete embeddings $X \in \domain{L\times d}$ into a continuous signal $\widetilde X(s): [0,1]\rightarrow \domain{d}$. This signal is expressed as a linear combination of $m$ radial basis functions (RBFs), denoted as $\Phi(s) \in \domain{m}$ with the coefficient matrix $B\in\domain{m\times d}$, which is fitted by multivariate ridge regression~\cite{brown1980adaptive}. This continuous signal representation allows for unbounded context representation with fixed memory storage, independent of the context length, similar to LSTM. However, as the memory cache is stored as a continuous signal, it cannot simply prepend to the current segment but has to transform back to embeddings via continuous attention~\cite{martins2020sparse}.
%
    \begin{align}
        & \widetilde O_{t}^N \defeq \mathrm{Transformer}(\widetilde X_{t}^0),\;\widetilde X_{t}^0 \defeq \left[ X_{t}^{mem} \circ X_{t}^0 \circ X_{t}^{mem} \right], \label{eq: rmt}\\
        &where\;\; X_{t}^{mem} \defeq O_{t-1}^{write},\; \left[O_{t-1}^{read}\circ O_{t-1}^{N}  \circ O_{t-1}^{write}\right] \defeq \widetilde  O_{t-1}^{N}\notag\\
        \notag\\
        & (\widetilde K^{N}_t,\widetilde V^{N}_t) \defeq \left[ \mathrm{retr}(Q^{N}_t,m,k) \circ (K^{N}_t,V^{N}_t) \right], \label{eq: knn_transformer}\\
        & where\;\; \mathrm{retr}(Q^{N}_t,m,k) \defeq \mathrm{kNN}\left[Q^{N}_t,\{(K^{N}_{t-\tau},V^{N}_{t-\tau})\}_{\tau=1}^{m}\right]\notag
    \end{align}

\noindent\textbf{Alternate Cache Designs}.
RMT~\cite{bulatov2022recurrent} formalizes the memory cache as special \texttt{[mem]} tokens, prepended both at the start and the end of each segment, as shown in Eq.~\ref{eq: rmt}. After processing each segment, the read/write tokens will be split from the output embeddings, and the write tokens will be taken as the \texttt{[mem]} tokens for the next segment. By leveraging such a recurrence mechanism with global memory tokens, RMT is demonstrated to scale effective context size to 1M tokens~\cite{bulatov2023scaling}. Memorizing Transformer~\cite{wu2022memorizing} applies (key, value) memory cache only for the top attention layer, but with a large cache size without compression. Besides, instead of a simple FIFO cache to read memory, they use \textit{k}NN algorithm to retrieve top-\textit{k} most similar (key,~value) pairs for each query to prepend to the local ones, as Eq.~\ref{eq: knn_transformer} indicates. In contrast, Memformer~\cite{wu2020memformer} reads and writes the memory cache fully leveraging variants of self-attention with a \textit{forgetting mechanism} to retrieve and retain the most significant information through long-range time steps.

\subsection{External MemoryBank}


The discussed mechanisms enhance the vanilla stateless Transformer with sequential recurrence by prepending additional hidden states from an internal memory cache. However, they present drawbacks. Firstly, minor changes in the memory mechanism may require full model retraining, underutilizing pretrained LLMs already proficient in capturing contextual dependencies, albeit not long enough. Secondly, as noted in Memorizing Transformer~\cite{wu2022memorizing}, they frequently face the challenge of \textit{memory staleness}, where older hidden states in the memory cache may diverge in distribution from the latest ones during training, thus hampering memory augmentation effectiveness.


As a solution, another retrieval-augmented mechanism, often named Retrieval-Augmented Generation (RAG)~\cite{lewis2020retrieval}, decouples the model from its long-term memory storage by using a contextual information encoder to store long sequences as segmented embedding vectors to an external memory bank. During queries, the model retrieves information from this memory bank based on specific criteria and dynamically concatenates it to form in-context working memory.


\noindent\textbf{Cosine-Based Retrieval Criteria}.
LangChain~\cite{langchain2022, topsakal2023creating} is an open-source framework tailored for applications like chatbots. It ingests user-specified local documentation in standard readable formats, vectorizing it via off-the-shelf LLMs into a memory bank. During interactions, it retrieves top-relevant contexts using dot-product \textit{cosine similarity} between user prompt embeddings and stored vectors, prepending them to the prompt for generating responses. This pipeline efficiently leverages off-the-shelf LLMs, enhancing their long-term memory with a cost-effective, flexible, and dynamic mechanism. Similar works have also developed external memory banks for chatbot applications~\cite{LangchainChatchat2023} and SQL generation~\cite{vanna_ai_vanna}.

\noindent\textbf{Heuristic Retrieval Criteria}.
Except for cosine similarity, RETRO~\cite{borgeaud2022improving} retrieves from the BERT-embedded KV memory bank via $k$NN search based on $L_2$ distance. Similarly utilizing $k$NN search, Unlimiformer~\cite{bertsch2023unlimiformer} enables any existing pretrained encoder-decoder transformer to index unlimited input sequences for decoder retrieval of top-$k$ keys for cross-attention. In contrast, SiliconFriend~\cite{zhong2023memorybank} proposes the \textit{Memory Bank} mechanism for tracking long-chat history, providing specialized responses through dialogue logging, event distillation, user personality awareness, and memory refreshment. RecurrentGPT~\cite{zhou2023recurrentgpt} facilitates recurrent prompting and defines the recurrent computation graph with ChatGPT by simulating long-/short-term memory mechanism in LSTM~\cite{yu2019review}. Moreover, RecallM~\cite{kynoch2023recallm} organizes and updates memory as a dynamic concept-aware knowledge graph for improved continual learning and temporal reasoning during chat. Inspired by \textit{Davidsonian semantics}~\cite{davidson1967logical}, Ret-LLM~\cite{modarressi2023ret} stores/writes and retrieves/reads knowledge as triplets $\langle A,B,R \rangle$ (each means "A and B have a relationship of R"), utilizing finetuned Alpaca~\cite{taori2023alpaca} to follow the instructions in memory read/write operations. Inspired by traditional operating systems, MemGPT~\cite{packer2024memgpt} implements a hierarchical memory management system, swapping contexts between "main memory" in the chat history and "disk storage" in the bank via function calls.

\noindent\textbf{Learnable Retrieval Criteria}.
Despite these heuristic designs, REALM~\cite{guu2020retrieval} pre-trains a latent \textit{neural knowledge retriever} using MLM as the learning signal, which takes charge of retrieving knowledge from a large textual corpus. LongMem~\cite{wang2023augmenting} trains another Transformer-based \textit{SideNet} to decouple the memory retrieval and fusion process from the pretrained LLMs, which are only responsible for encoding the (key, value) pairs into the memory bank. Recently, FOT~\cite{tworkowski2023focused} proposes a novel contrast training procedure across batches of documents, reshaping the KV space to address \textit{distraction issue} as the size of a $k$NN-lookup memory bank increases during inference.

\section{Extrapolative PEs}\label{subsec: extrapolative_pe}

Recognizing the need to push the inference length boundary beyond $L_{max}$, the research community has made significant efforts in this direction. Notably, according to ~\cite{anil2022exploring}, they have determined that \textit{distractors} are the primary cause of failures in length generalization in the case of \textit{parity} task. These issues, however, can be mitigated considerably through approaches such as scratchpad prompting~\cite{nye2021show}. Nevertheless, in this section, our focus remains on the undeniable role that current PEs play in length generalization in more general scenarios.

\subsection{Enhancing Understanding}\label{subsubsec: pe_insights}

Before entering the concrete approaches, we would love to provide some insights below to enhance the understanding of this minute but essential design in Transformer for sequential modeling tasks.

\noindent\textbf{Rethinking PEs as $\beta$-Encoding}.
Su~\cite{transformer-upgrade-10} revisits the sine and cosine basis functions of SinPE and RoPE, considering them as approximated terms for the $\beta$-encoding system to represent any position number $n$, as shown in Eq.~\ref{eq: beta_encoding}. This approach employs $\frac{d}{2}$ fixed $\beta$-bits, where $\beta \defeq \theta^{-1} = base^{2/d}$ represents the power basis of the wavelength or period of the trigonometric basis functions, which increases as a geometric series $\{\beta^i\}_{i=0}^{d/2}$ with the dimension $i$ goes deeper.

To gain a deeper understanding of this concept, we can draw a comparison between Eq.~\ref{eq: beta_encoding} and Eq.~\ref{eq: sinpe}-\ref{eq: rope}. It becomes evident that the $i$-th $\beta$-bit of the representation of $n$ involves the division of the $i$-th power of $\beta$, followed by some sort of periodic operations ($\mod$ in Eq.~\ref{eq: beta_encoding} and $\sin, \cos$ in Eq.~\ref{eq: sinpe}, \ref{eq: rope}).
    \begin{align}
        & n(\beta) \defeq \left\{\; \lfloor \cfrac{n}{\beta^{i}} \rfloor \;\mathrm{mod}\; \beta \;\right\}_{i=0}^{\lceil \log_{\beta} n \rceil-1}  \label{eq: beta_encoding}
    \end{align}

\noindent\textbf{Length Extrapolation Dilemma}.
Before the era of Transformers, RNN-based language models were trained on shorter sequences but were expected to generalize effectively to longer contexts, a phenomenon referred to as \textit{length extrapolation} or \textit{length generalization}~\cite{mikolov2010recurrent, mikolov2012context}. Unfortunately, recent studies~\cite{press2021train,chen2023extending, anil2022exploring} have highlighted a significant shortcoming of \textit{length extrapolation} ability for Transformer-based language models. This causes the insufficient context length limit during inference when applying to real-world applications, as analyzed in Sec.~\ref{subsec: limitation_analyses}.

In the original Transformer paper~\cite{vaswani2017attention}, there is little discussion regarding the design insights or theoretical interpretation of their SinPE. This has led many researchers to question its necessity and effectiveness, especially the blame on the extrapolation deficit, which also points to the same trigonometry-based RoPE~\cite{su2021roformer}. To understand the lousy extrapolation caused by current trigonometric PEs, we investigate and summarize two insights from distinct views as follows. 

\begin{itemize}
\item From a mathematical view, as Su~\cite{transformer-upgrade-7} explains in his blog, extrapolation, which involves inferring the whole from local information, depends on the high-order \textit{smoothness} of the function. However, these PEs are designed as combinations of high-frequency oscillatory trigonometric basis functions to accommodate sufficient positional information. This choice makes it challenging for the models to generalize without specific learning during training stages. 
    
\item From a training view, due to the wavelength or period of the basis functions increasing exponentially, proportional to $\{\beta^i\}_{i=0}^{d/2}$, training samples constrained by currently supported $L_{max}$ are typically too short for the rear low-frequency dimensions to span the entire periodic cycle. This suggests only a few dimensions perceive complete periodic information, thus receiving sufficient training for extrapolation, and the boundary is defined as \emph{critical dimension} in ~\cite{liu2023scaling} (e.g., for Llama2-4k~\cite{touvron2023llama2}, the critical dimension is only 92). Consequently, direct extrapolation becomes prone to failure when relying on these poor-learned low-frequency components.
\end{itemize}

\subsection{Attention Bias}
As alternative mechanisms to explicitly encoding positional information, \textit{attention bias} have been explored to capture the sequentiality and temporality of natural language incorporated into the attention kernel. As shown in Eq.~\ref{eq: attention_bias}, the attention bias is depicted as a matrix, denoted as $B$, added to the unnormalized attention weights matrix $P$ before applying the softmax operation. Each element of this matrix, indexed by $(i,j)$, carries positional information encoded by a function $\mathcal{B}$. Thus, it is reasonable to regard the attention bias as a form of relative PEs.
%
    \begin{align}
        & \widetilde{P} \defeq P + B, \quad B \in \domain{L\times L},\quad where\;\; B_{ij} \defeq \mathcal{B}(i,j), \;\;\forall i,j \in \{0,1,\ldots ,L-1\}\label{eq: attention_bias}
    \end{align}

Early approaches like T5~\cite{raffel2020exploring} employ learnable attention bias, denoted as $\mathcal{B}_{\theta}(i,j)$, which is independent for each head in each attention layer. However, they did not explicitly address the problem of length extrapolation. The breakthrough in recognizing and addressing the extrapolation problem comes with ALiBi~\cite{press2021train}. ALiBi introduces a negative causal attention bias heuristically, as shown in Eq.~\ref{eq: alibi}, where $\lambda^{(h)}$ is a head-specific slope fixed before training and decreases geometrically with the head index $h$. ALiBi successfully maintains low perplexity levels when extrapolating inference tokens beyond $L_{max}$ up to 16$\times$.

Following the success of ALiBi, several variants emerged in the quest to improve extrapolative PEs for Transformer-based LLMs. KERPLE~\cite{chi2022kerple} extended the ALiBi-style attention bias by considering it as a composition triangle kernel to self-attention. Two extra learnable scalar parameters were introduced to generalize the bias kernel, as shown in Eq.~\ref{eq: kerple}. The authors of Sandwich~\cite{chi2023dissecting} reused the Sinusoidal PEs to form the attention bias in a RoPE-style, as illustrated in Eq.~\ref{eq: sandwitch}, with $\lambda$ as a hyper-parameter to tune. Interestingly, another method discussed by Su~\cite{transformer-upgrade-7} in his blog utilizes a \textit{super-baseline} approach during inference, as illustrated in Eq.~\ref{eq: super_baseline}. This method relies on a local causal attention mask, where each query attends to keys whose distances have not exceeded $L_{max}$ while still applying RoPE. According to Su's experiments, this approach proves to be simple, low-cost, and performs sufficiently well compared to the more elaborate designs mentioned earlier, thus referred to as a \textit{super-baseline}.
%
    \begin{align}
        & \mathcal{B}_{ALiBi}^{(h)}(i,j) \defeq -\lambda^{(h)}\cdot |i-j|, \;\;\lambda^{(h)} \defeq \cfrac{1}{2^h} \;or\; \cfrac{1}{2^{h/2}}\label{eq: alibi}\\
        & \mathcal{B}_{\textit{KERPLE}}(i,j) \defeq \begin{cases}
            -r_1\log(1 + r_2|i-j|), \quad r_1, r_2 > 0\\
            -r_1 |i-j|^{r_2}, \quad r_1 > 0, r_2 \in (0,2]\\
        \end{cases}\label{eq: kerple}\\
        & \mathcal{B}_{Sandwitch}(i,j) \defeq \lambda\cdot \langle \mathrm{SinPE}(i), \mathrm{SinPE}(j) \rangle\label{eq: sandwitch}\\
        & \mathcal{B}_{super\text{-}baseline}(i,j) \defeq \begin{cases}
            0, & |i-j| \in [0, max\text{-}length]\\
            -\infty, & otherwise\\
        \end{cases}\label{eq: super_baseline}
    \end{align}

\subsection{Extended RoPE}\label{subsubsec: extended_rope}
RoPE, as introduced in Sec.~\ref{subsec: preliminaries}, is a widely-used positional encoding scheme utilized in popular LLMs such as Llama and GLM. It offers advantages such as relative distance decay, training stability, compatibility with linear attention, and better length extrapolation capabilities compared to the traditional SinPE, as demonstrated in various experiments ~\cite{press2021train,chen2023extending}, albeit not that satisfactory. Therefore, several research works have aimed to extend RoPE using various strategies to enhance its length extrapolation capabilities. 


\noindent\textbf{Scaling Strategies}.
Recent approaches in the community have gained prominence by scaling to extrapolate inference context length with minimal or no finetuning~\cite{chen2023extending, ntk-aware-rope}. Aside from simple modifications to the \textit{base} parameter~\cite{xiong2023effective, ntk-aware-rope, roziere2023code}, LEX~\cite{sun2022length} introduces XPOS, an extended causal RoPE incorporating an additional exponential decay term, represented by Eq.~\ref{eq: lex_xpos}, where $\gamma \in (0,1)$ is a scalar hyper-parameter. Similar techniques are utilized in PermuteFormer~\cite{chen2021permuteformer} to adapt to Performer~\cite{choromanski2020rethinking}. Additionally, Positional Interpolation (PI)~\cite{chen2023extending} employs linear scaling on each position number from $n$ to $\frac{n}{\kappa}$, densifying the representation space to extend the length boundary by $\kappa$ times (cf. Eq.~\ref{eq: pi}). This strategy proves experimentally more stable and requires fewer finetuning steps than direct extrapolation.


However, linear scaling may hinder the network's ability to distinguish the order and positions of closely spaced tokens, compressing their distances by a ratio of $\kappa$. Drawing from the Neural Tangent Kernel theory (NTK)~\cite{jacot2018neural}, which suggests that deep neural networks struggle with learning high-frequency information when input dimension is low and corresponding embeddings lack high-frequency components, NTK-aware Scaling RoPE (NTK-RoPE)~\cite{ntk-aware-rope} combines high-frequency extrapolation and low-frequency interpolation. It scales $\beta$ using coefficient $c_{\kappa}$ to achieve equivalence during interpolation by a ratio of $\kappa$ for the lowest frequency term while maintaining scale for terms with higher frequency (see Eq.~\ref{eq: ntk}). Surprisingly, this nonlinear scaling can be directly applied to LLMs pretrained with RoPE, like Llama, without further finetuning to extend the context length boundary, adopted in CodeLlama~\cite{roziere2023code}.

Inspired by NTK-RoPE, several enhanced scaling methods have emerged. To avoid performance degradation when $L$ is still within the $L_{max}$, Dynamic-NTK~\cite{dynamic-ntk, peng2023yarn} delays applying the NTK scaling trick until $L$ exceeds the supported context length $L_{max}$, gradually increasing ratio $\kappa$ as $L$ grows, adopted in Qwen-7B~\cite{qwen-7b} and updated Llama2~\cite{llama2-dynamic-ntk}. To generalize $\beta$-scaling across dimensions, NTK-mix RoPE~\cite{transformer-upgrade-11} introduces multiple coefficients $c_{\kappa}^{(0)}\ge c_{\kappa}^{(1)}\ge \ldots \ge c_{\kappa}^{(d/2-1)}$ for $\beta$ to interpolate less as the frequency increases. In contrast, NTK-by-parts~\cite{ntk-by-parts, peng2023yarn} avoids interpolating higher frequency dimensions while always interpolating lower ones. YaRN~\cite{peng2023yarn}, an extension of NTK-RoPE, combines NTK-by-parts with a \textit{length scaling} trick that scales $Q, K$ by a constant temperature factor $t$. Giraffe~\cite{pal2023giraffe} introduces \textit{Power Scaling} (Eq.~\ref{eq: giraffe_power_scaling}), where the exponent $\kappa>0$ controls the decay ratio of low frequencies, ensuring high-frequency elements are less affected than poorly learned low-frequency ones. Despite manual designs, CLEX~\cite{chen2023clex} employs a neural ordinary differential equation (ODE) to learn continuous scaling as a dynamical system.

    \begin{align}
        \mathrm{XPOS}:&\;\;  P_{i,j} \defeq \langle\widetilde{\bold q}_i, \widetilde{\bold k}_j \rangle = \gamma^{i-j}(\bold q\transpose R_{j-i} \bold k),
        \;\;\text{where}\;\; \widetilde{\bold q}_i \defeq \gamma^i(R_i \bold q), \; \widetilde{\bold k}_j \defeq \gamma^{-j} (R_j \bold k),\; i \ge j\label{eq: lex_xpos}\\
        \mathrm{PI}:&\;\;  \widetilde P_{i,j} \defeq \langle R_{i/\kappa}\bold q, R_{j/\kappa}\bold k\rangle = \bold q^{\mathrm{T}} R_{\frac{j-i}{\kappa}} \bold k \label{eq: pi}\\
        \mathrm{NTK}:&\;\;  \widetilde\beta \defeq c_{\kappa}\cdot\beta, \;\;s.t.\;\;\cfrac{n}{\widetilde\beta^{d/2-1}} = \cfrac{n/\kappa}{\beta^{d/2-1}} \Rightarrow c_{\kappa} = \kappa^{2/(d-2)} \label{eq: ntk}\\
        \mathrm{Power}\; \mathrm{Scaling}:&\;\; \widetilde\beta^{i} \defeq \beta^{i} / (1-2i/d)^{\kappa} \label{eq: giraffe_power_scaling}
    \end{align}

\noindent\textbf{Truncation Strategies}. 
Based upon the idea of \textit{high-frequency extrapolation and low-frequency interpolation} from NTK-RoPE, Su further proposes two simple truncation strategies in his blog~\cite{transformer-upgrade-12}, named \textit{ReRoPE} and \textit{Leaky ReRoPE} after the activation function Rectified Linear Unit (ReLU) and its leaky variant. As shown in Eq.~\ref{eq: rerope}, the main idea behind this \textit{Rectified Truncation} approach is to set a local window with size $w$, and for each token, no scaling is applied as long as the tokens attend are inside the window. However, linear scaling, akin to Leaky ReLU, increases the position by step $1/\kappa$ when the token is located outside the window (Leaky-ReRoPE). This method combines high-frequency extrapolation and low-frequency interpolation more directly and ensures that $L_{max}$ is not exceeded by carefully tuning $w$ and $\kappa$. Furthermore, suppose the ratio $\kappa$ is set to infinity. In that case, it applies a constant PE of position number $w$ to any pair of $(\bold q_i, \bold k_j)$ as long as $|i-j|\ge w$, potentially accommodating infinite contexts (ReRoPE). According to Su's and our elementary experiments, ReRoPE performs very well without finetuning on perplexity metric and QA tasks, even outperforming NTK-based schemes.


However, Leaky ReRoPE and ReRoPE entail two stages of scaling without a linear transformation to bridge their gap. Consequently, they require two attention matrix computations per stage and use a boolean matrix to merge them, significantly increasing inference cost and limiting the effective length boundary. Moreover, they are currently incompatible with Flash Attention~\cite{dao2022flashattention, dao2023flashattention} to mitigate high computational costs. To adapt ReRoPE with Flash Attention, we have re-implemented the Flash Attention forward kernel to incorporate ReRoPE based on Triton~\cite{tillet2019triton}, somewhat alleviating its computational burden.\footnote{The experimental implementation is available at: \url{https://github.com/Strivin0311/long-llms-learning/blob/main/notebooks/flash_rerope.py}.} Additionally, Giraffe~\cite{pal2023giraffe} introduces another truncation strategy, called \textit{Basis Truncation}, depicted in Eq.~\ref{eq: giraffe_basis_truncation}, where $a,b$ are cutoff thresholds. This approach retains high-frequency basis components while reducing low-frequency elements to near-zero values ($\rho \approx 0$), simplifying extrapolation for low-frequency components.

%
    \begin{align}
        \mathrm{Rectified\; Truncation}:&\;\; \widetilde P_{i,j} \defeq \langle R_{\alpha(i,j,w,\kappa)} \bold q,\;\bold k\rangle, \label{eq: rerope} \text{where}\\
        &    \alpha(i,j,w,\kappa) \defeq \begin{cases}
          \min\{ i-j, w+\frac{i-j-w}{\kappa}\}, & 0<\kappa<\infty\; (\mathrm{Leaky\; ReRoPE})\\
          \min\{i-j,w\} & \kappa \rightarrow \infty\; (\mathrm{ReRoPE})
        \end{cases}\notag\\
        \mathrm{Basis}\; \mathrm{Truncation}:&\;\; \widetilde\theta^{i} \defeq \begin{cases}
        \theta^{i}, & \theta^{i} \ge b\\
        \rho, & \theta^{i} \in (a,b) \\
        0, & \theta^{i} \le a
        \end{cases} \label{eq: giraffe_basis_truncation}
    \end{align}

\noindent\textbf{Rearrangement Strategies}.

Based on the insights from Sec.~\ref{subsubsec: pe_insights}, it is evident that rear position embeddings are updated less frequently than front ones, potentially leading to improperly trained rear positions. Recent works address this issue effectively. SHAPE~\cite{kiyono2021shape} achieves shift invariance by randomly shifting absolute positions during training. Random Padding~\cite{tao2023frustratingly} balances updating times across all positions by moving an arbitrary number of padding tokens to the front during finetuning. Randomized PE~\cite{ruoss2023randomized} trains with a randomly sub-sampled set of positions from a broader range than the sequence length, enhancing robustness. PoSE~\cite{zhu2023pose} finetunes models to adapt all relative positions of the target context window by adding a distinct \textit{skipping bias term} to position indices of training samples to simulate longer inputs.

In summary, research on extrapolative PEs is a promising and rapidly developing field, aiming to enhance the LLMs' ability to infer long contexts in real-world scenarios with an available $L_{max}$ setting during training.

\section{Context Processing}\label{subsec: context_processing}

Many methods propose intricate designs around the attention module in the Transformer architecture. In contrast, there exist simpler approaches that treat pretrained LLMs as black-box or gray-box models and handle long-context inputs by making multiple calls to the model, ensuring that each call respects the $L_{max}$ limitation. While these approaches don't enhance the LLMs' inherent ability to process long contexts, they leverage the models' in-context learning capabilities, albeit with increased computation and potentially less precise answers.

\subsection{Context Selection}

To fit long segments within the context window of LLMs while preserving relevant information, some works partition lengthy texts into segments and select specific ones based on predefined criteria. They vary in defining selection criteria with corresponding scores, either sorting simultaneously or picking iteratively.


LangChain~\cite{langchain2022} employs three strategies for handling retrieved context that exceeds $L_{max}$, one of which is referred to \textit{Map Rerank}, querying LLMs to output answers and confidence scores independently for each segment, and select the answer with the highest confidence score as the final output. CogLTX~\cite{ding2020cogltx} introduces a multi-step reasoning mechanism, \textit{MemRecall}, where two models sequentially score context segments in a coarse-to-fine manner during each reasoning step. The top-\textit{k} segments with the highest scores are added to the final candidate queue, deferring the remaining segments to the next reasoning step until the queue is filled. LoBART~\cite{manakul2021long} uses ROUGE-2 score~\cite{ganesan2018rouge} to select the top-\textit{k} contexts during training and trains an additional Hierarchical RNN model~\cite{cho2014learning, cohan2018discourse} for generating surrogate priority scores for context selection during inference.

\subsection{Context Aggregation}

In contrast to selection-based methods, some approaches consider contributions from all context segments to the final answer rather than selecting one. Initially, they extract relevant information from each segment individually and then employ various fusion strategies to aggregate the retrieved information, arriving at the final answer. These approaches vary in two key aspects: how information is extracted from each segment and the fusion strategies used to integrate information across all segments.


\noindent\textbf{Fusion in Decoder}.
For LLMs like T5 and BART, a category of methods known as Fusion-in-Decoder (FiD)~\cite{izacard2020leveraging} leverages the Encoder to extract information as embedded hidden states and the Decoder to attend to all contextualized representations to generate the final output. In SLED~\cite{ivgi2023efficient}, for example, contextualized representations are created by overlapping a small portion of each segment with neighboring segments, forming \textit{context paddings}. Each segment is then independently encoded through the Encoder, and the resulting embeddings are concatenated to generate embeddings for the entire extended document, excluding the context paddings. Finally, the Decoder integrates these locally contextualized embeddings through cross-attention, achieving coherent fusion of information.


\noindent\textbf{Map Reduce and Refinement}. LangChain~\cite{langchain2022} introduces two additional aggregation techniques. The first, \textit{Map Reduce}, involves processing each segment simultaneously to obtain answers in parallel. These answers are then synthesized into a final summary by another LLM. In contrast, the second approach \textit{Refine} progressively refines answers throughout the processing of each segment. Answers from previous segments are cascaded with the current segment, serving as prompts for further refinement. This iterative process continues until the final segment is processed.


\noindent\textbf{Parallel Context Windows}. PCW~\cite{ratner2023parallel, hao2022structured} handles long-context inputs similarly. It partitions the extended context into multiple smaller windows, each with a maximum length $C \defeq L - T$, where $L$ is the total context length and $T$ is the length of task-related tokens in the query, along with the maximum number of new tokens to be generated. Within each window, tokens attend to each other in parallel, with their position indices isolated within the range of $[0, C-1]$. Task-related tokens, including the last context position indices within the range $[C, L-1]$, attend to all context tokens, aggregating parallel information from each window to generate the final answer.


Similarly, NBCE~\cite{nbce} treats parallel context windows as a series of independent conditions $C_1, C_2, \ldots, C_n$, approximating the logarithmic posterior probability $\log \mathrm{P}(T|C_1, C_2, \ldots, C_n)$ for token generation. It leverages the \textit{Naive Bayes} algorithm~\cite{webb2010naive} to simplify the formulation. As deduced in Eq.~\ref{eq: nbce}, $\mathrm{P}(T|C_i)$ represents the likelihood conditioned on the $i$-th context window, $\mathrm{P}(T)$ represents the prior, and \textit{const} is a constant depending solely on $C_1, C_2, \ldots, C_n$. Su extends this formulation to a more general case (cf.\ Eq.~\ref{eq: nbce_param}), introducing a hyper-parameter $\mu$ and a pooling operation, denoted as \textit{pool}, which can be either \textit{average} or \textit{max}. In this extended form, the original \textit{Naive Bayes} formula is a specific instance where $\mu$ is set to $n-1$ and \textit{pool} is defined as \textit{average}.
%
    \begin{align}
        &\log \mathrm{P}(T|C_1,\ldots ,C_n) \xlongequal[NB]{} \log \left[ \frac{\mathrm{P}(T)\prod\limits_{i=1}^n \frac{\mathrm{P}(T|C_i)\mathrm{P}(C_i)}{\mathrm{P}(T)}}{\mathrm{P}(C_1,\ldots,C_n)} \right] = \sum\limits_{i=1}^n \log \mathrm{P}(T|C_i) - (n-1) \log \mathrm{P}(T) + const \label{eq: nbce}\\
        & \Rightarrow (\mu+1) \mathrm{pool}[\log \mathrm{P}(T|C_i)] - \mu \log \mathrm{P}(T) + const \label{eq: nbce_param}
    \end{align}


NBCE and PCW can be applied to any readily available LLMs but assume negligible relationships among context windows, treating them uniformly and unordered. Their performance may suffer with tightly interconnected or excessive windows to process in parallel.

\subsection{Context Compression}

Despite the selection or aggregation of source long contexts, some works focus on directly compressing the (hidden) sequence length under the $L_{max}$ constraint. They aim to produce more condensed and abstract representations of the long raw contexts before feeding them into LLMs, through either learning embedded alternatives (\textit{soft compression}) or filtering out redundancies (\textit{hard compression}) based on various scores computed by pretrained models.


\noindent\textbf{Soft Compression}.
\cite{wingate2022prompt} optimizes a few soft prompt tokens to significantly compress the original prompt while retaining abstract sentiments. To avoid optimizing for every new context, AutoCompressor~\cite{chevalier2023adapting} compresses embedded long segmented contexts into \textit{summary vectors} by training an RMT-based pretrained model~\cite{bulatov2022recurrent} using a simple unsupervised objective. For instruction-following, \cite{mu2024learning} learns a gist model to compress instructions into prefixed \textit{gist tokens}, which can be reused for trained tasks and also generalized to novel ones.


\noindent\textbf{Hard Compression}.
\cite{li2023compressing} utilizes self-information values~\cite{bunescu2022distribution} to filter out redundant or non-essential tokens. LLMLingua~\cite{jiang2023llmlingua} employs a coarse-to-fine compression process to rephrase various components in prompts, such as instructions and demonstrations, based on perplexity values. \cite{fei2023extending} initially segments long contexts into topic-based chunks using graph representation, followed by summarizing semantic-relevant sentences within each chunk.

\section{Miscellaneous}\label{subsec: miscellaneous}


This section provides a concise overview of miscellaneous solutions that extend the previously discussed four categories, offering a broader perspective on enhancing the effective context window of LLMs or optimizing the efficiency when using off-the-shelf LLMs. It is worth noting that the literature covered here may not be exhaustive or specific to Transformer-based models. Many of these techniques are applicable universally to any model equipped with deep neural networks, albeit particularly crucial for large-scale LLMs.


\noindent\textbf{Specific Objectives}.
In contrast to conventional pretraining objectives like MLM or CLM (discussed in Sec.~\ref{subsec: preliminaries}), recent research explores tailored approaches to adapt pretraining for specific tasks, aiming to enhance LLMs' effectiveness in capturing intricate long-range dependencies and discourse structures in longer texts compared to shorter ones~\cite{dong2023survey}. 
For instance, XLNet~\cite{yang2019xlnet} introduces a permutation objective that excels in various NLP tasks. ERNIE-Doc~\cite{ding2020ernie} extends this approach to long documents with a \textit{Segment-Reordering Objective} to model long-range relationships. DANCE~\cite{gidiotis2020divide} employs a \textit{divide-and-conquer} preprocessing strategy for summarization tasks, breaking the long document and its summary into multiple source-target pairs. PEGASUS~\cite{zhang2020pegasus} introduces the Gap Sentence Generation (GSG) objective for abstractive summarization, while PRIMERA~\cite{xiao2021primera} extends it across multi-documents using the \textit{Entity Pyramid} method.


\noindent\textbf{Mixture of Experts}. MoE~\cite{jacobs1991adaptive, shazeer2017outrageously, openmoe2023} augments giant LLMs by replacing the dense FFN layer with a MoE layer, incorporating multiple specialized \textit{experts} where each excels in specific input types or tasks. A dynamic \textit{gating mechanism} selects the most suitable expert for a given input, which can be implemented through various ways, including task-optimized expert modules~\cite{jacobs1991adaptive, chen2023adamv}, sparse activation~\cite{shazeer2017outrageously, jaszczur2021sparse, gupta2022sparsely}, sharding across multiple devices~\cite{lepikhin2020gshard, he2022fastermoe}, and adapting mixture weights through training to determine the contribution of each expert~\cite{puigcerver2023sparse}. Then, routing mechanisms select the top-\textit{k} experts for each token based on their gate values. \cite{shazeer2017outrageously} sets $k>1$ to determine the number of experts, while Switch Transformer~\cite{fedus2022switch} suggests $k=1$ to both preserve model quality and reduce routing computation, known as \textit{Switch Routing}. Finally, the output is obtained through a weighted summation of contributions from selected experts~\cite{shazeer2017outrageously}. MoE techniques significantly enhance versatility, reduce computational demands, and elevate the efficiency and effectiveness of modeling large-scale contexts~\cite{csordas2023switchhead, santos2023memory, li2023accelerating, nie2023flexmoe}, already adopted by Mixtral~\cite{mixtral2023mistralai}.

\noindent\textbf{Parallelism}. 
Leveraging modern aggregated GPU memory within and across nodes, recent research has introduced various parallelism strategies to scale up model sizes and extend sequence length. We summarize commonly used parallelism paradigms with brief introductions as follows.

\begin{itemize} 
    \item[(a)] \textit{Data Parallelism (DP)}~\cite{li2020pytorch}, widely integrated into PyTorch~\cite{pytorch-data-parallelism}, is the most commonly used way to accelerate training in a distributed manner across multiple devices. It replicates the model on each device to generate gradients independently and communicates them at each iteration to maintain consistency.
    \item[(b)] \textit{Tensor Parallelism (TP)}~\cite{shoeybi2019megatron} introduces tensor splitting, where individual layers of the model are horizontally partitioned over multiple devices.
    \item[(c)] \textit{Pipeline Parallelism (PP)}~\cite{huang2019gpipe, harlap2018pipedream} splits the model layers vertically along the batch dimension into different partitions of micro-batches on separate devices. Each device processes one micro-batch received from the previous one in a pipeline fashion.
    \item[(d)] \textit{Sequence Parallelism (SP)}~\cite{li2021sequence, korthikanti2022reducing} divides the input sequence into multiple chunks and feeds each chunk into its corresponding device. It incorporates ring-style communication for computing the attention output.
    \item[(e)] \textit{Expert Parallelism (EP)}~\cite{fedus2022switch}, as discussed earlier in MoE, places different experts on different GPUs and executes them in parallel. Classic \textit{all-to-all} communication primitives are often used to implement this form of parallelism~\cite{rajbhandari2022deepspeed}.
\end{itemize}

Furthermore, the integration of various parallelism strategies are prevailing in distributed environments, including \textit{3D parallelism} by DeepSpeed~\cite{team2020deepspeed}, which combines (a)--(c) with ZeRO~\cite{rajbhandari2020zero, wang2023zero++}, AutoParallelism by Colossal-AI~\cite{li2023colossal}, FSDP by PyTorch~\cite{zhao2023pytorch}, and promising \textit{4D parallelism}~\cite{li2021sequence4d, pipegoose} by adopting (d) into existing mechanisms to both scale parameters as well as sequence length.


\noindent\textbf{Weight Compression}. 
Various methods enhance memory efficiency in large-scale LLMs through weight compression techniques, including pruning~\cite{michel2019sixteen, sun2023simple, ma2024llm}, factorization~\cite{lan2019albert}, quantization~\cite{zafrir2019q8bert}, partitioning~\cite{pope2023efficiently}, and distillation~\cite{wang2020minilm}. Among them, quantization strategies particularly play a crucial role in practical deployment of massive LLMs, by reducing parameter precision~\cite{dettmers2022llmint8, dettmers20228bit, kuzmin2022fp8}, thereby alleviating memory demands and accelerating both training~\cite{dettmers2023qlora, xu2023qa, li2023loftq, kim2024memory} and inference~\cite{lin2023awq, frantar2023gptq, badri2023hqq}. Additionally, simpler approaches exist to mitigate the large KV cache during inference, such as Multi-Query Attention (MQA)~\cite{shazeer2019fast} and Grouped-Query Attention (GQA)~\cite{ainslie2023gqa}, which save the number of heads for KV and distribute them equally across multiple queries, applied to GLM and PaLM.

    \section{Evaluation Necessity \& Optimization Toolkit}\label{sec: necessity_toolkit}


In this section, we explore evaluation necessities for assessing long-context capabilities of LLMs, including datasets, metrics, and baseline models. Additionally, we investigate popular optimization toolkits, such as libraries, frameworks, and compilers, to enhance LLM efficiency and effectiveness during development. Detailed information is organized in table-format in Appendix~\ref{appendix: datasets}, \ref{appendix: metrics}, \ref{appendix: baselines}, and \ref{appendix: toolkits}, with a concise overview as follows.



\noindent\textbf{Datasets}.
We provide a collection of evaluation datasets commonly used to assess the long-context capabilities of LLMs, including benchmark suites like LRA~\cite{tay2021long}, SCROLLS~\cite{shaham-etal-2022-scrolls}, LEval~\cite{an2023leval}, LongBench~\cite{bai2023longbench}, InfiniteBench~\cite{zhang2023infinitebench}, and other single-task ones. In Tab.~\ref{tab: datasets} of Appendix~\ref{appendix: datasets}, detailed information on each dataset is available, covering language, task types, length statistics, quality, splits, and more. To ensure ongoing relevance, we maintain an updated version of this table on our GitHub repository, accessible via the following link: \url{https://github.com/Strivin0311/long-llms-learning/blob/main/evaluation/datasets.md}.


\noindent\textbf{Metrics}. 
We offer a summary of nine categories of general evaluation metrics commonly employed across ten NLP task types, encompassing language modeling, question answering, summarization, math solving, code generation, and open-ended writing, among others. For detailed metrics specific to each task, readers can refer to Tab.~\ref{tab: metrics} in Appendix~\ref{appendix: metrics}. Also, an accessible version of these metrics is available on our github repository at \url{https://github.com/Strivin0311/long-llms-learning/blob/main/evaluation/metrics.md}.


\noindent\textbf{Baselines}.
We gather a list of pretrained/finetuned LLMs commonly referenced in the literature, serving as baselines for evaluating long-context capabilities across various downstream tasks, such as Claude2 and GPT4 (close-sourced), along with LLongMA~\cite{peng2023yarn} and LongChat~\cite{longchat2023} (open-sourced). In Tab.~\ref{tab: baselines} of Appendix~\ref{appendix: baselines}, we present an overview of these models, including their basic information (e.g. $L_{max}$), statistics (e.g. parameter size, memory occupancy), and relevant links (e.g. Hugging Face, GitHub, blog / paper). For updates on the latest state-of-the-art baselines, please refer to \url{https://github.com/Strivin0311/long-llms-learning/blob/main/evaluation/baselines.md}.


\noindent\textbf{Toolkit}.
We collect a diverse array of valuable toolkits at Tab.~\ref{tab: toolkits} in Appendix~\ref{appendix: toolkits}, including libraries like vLLM~\cite{kwon2023efficient}, compilers like Triton~\cite{tillet2019triton} and frameworks such as DeepSpeed~\cite{deepspeed2020github} and Megatron-LM~\cite{shoeybi2019megatron}, to optimize the efficiency and effectiveness of LLMs across their development lifecycle. Updates and additional details can be accessed via \url{https://github.com/Strivin0311/long-llms-learning/blob/main/toolkits}.
    
    \section{Discussion}\label{sec: discussion}


Considerable progress has been achieved as we discussed in Sections \ref{subsec: effcient_attention}, \ref{subsec: long-term_memory}, \ref{subsec: extrapolative_pe}, \ref{subsec: context_processing}, yet several challenges persist. In this section, we explore these key challenges and suggest potential avenues for future research and development aimed at enhancing long-context capabilities of LLMs, especially architectural advancements for Transformers.


\noindent\textbf{Attention Trade-off}. 
As discussed in Section~\ref{subsec: effcient_attention}, efficient attention methods involve a trade-off between maintaining full-scale attention dependencies and achieving higher attention score precision to mitigate computational demands. With longer contexts, capturing global dependencies while preserving relevancy becomes crucial. Balancing computational efficiency with attention precision remains a key challenge in long-context LLMs. Recent innovations like Flash Attention~\cite{dao2022flashattention, dao2023flashattention} offer IO-aware solutions, significantly improving efficiency in runtime and memory usage without sacrificing attention precision. Integrating these solutions with existing strategies, and fuse them to GPU kernels with tools like Triton~\cite{tillet2019triton} (cf.\ SCFA~\cite{pagliardini2023faster}), presents promising avenues for practical application.


\noindent\textbf{Memory Efficacy and Efficiency}. 
As outlined in Sec.~\ref{subsec: preliminaries}, ~\ref{subsec: limitation_analyses}, we have identified limitations stemming from the absence of explicit memory mechanisms, relying solely on in-context working memory, and the significant increase in KV cache memory consumption during extended context interactions. These challenges emphasize the need for more effective and efficient memory mechanisms in Transformer-based LLMs. The long-term memory mechanisms discussed in Section~\ref{subsec: long-term_memory} face constraints due to additional memory overhead from intricate heuristic design, potentially leading to performance degradation over time. To address this, researchers explore more efficient strategies for memory organization and read/write throughput enhancement, drawing on recent advancements like Paged Attention~\cite{kwon2023efficient}.


\noindent\textbf{Length Extrapolation Mining}. 
In Section \ref{subsec: extrapolative_pe}, we analyze challenges in length extrapolation in Transformer-based models, focusing on positional embeddings. And we overview recent breakthroughs, including extended strategies applied to RoPE~\cite{ntk-aware-rope, dynamic-ntk, ntk-by-parts, peng2023yarn, transformer-upgrade-12, chen2023clex}, which show promise in addressing this limitation. However, these advancements often rely on simplified observations of positional embedding properties and heuristic adjustments. This prompts us to question the theoretical foundations of modeling sequentiality with high-dimensional embeddings and explore the potential resurgence of learnable embeddings with many hyper-parameters. Future research, exemplified in CLEX~\cite{chen2023clex}, should delve deeper into this area, especially in developing a robust theoretical framework for modeling sequentiality in Transformer settings.


\noindent\textbf{Specific yet Universal Objective}. 
While we have discussed objectives tailored for long-text modeling, many are limited to certain tasks or compatible only with the MLM objective rather than the more common CLM objective nowadays. This underscores the need for universally applicable causal language modeling objectives that can capture long-range dependencies effectively from the early stages of training. Aligning such objectives with an effective PE scheme, as mentioned earlier, could achieve this.


\noindent\textbf{Reliable Metric Demand}.
In Section~\ref{sec: necessity_toolkit}, we explored various evaluation metrics. However, our experience highlights significant disparities between commonly used metrics and human judgment~\cite{kuribayashi2021lower}. With LLMs rapidly deployed in real-world scenarios, there is a growing need for more dependable metrics~\cite{meister2021language}, especially in generative tasks where precise ground truth is elusive. One promising approach is leveraging state-of-the-art LLMs like GPT4 as substitutes for human judges, though high associated costs challenge wider adoption in the research community.


    \section{Conclusion}\label{sec: conclusion}

In this survey, we comprehensively navigate the landscape of architectural advancement in Transformer-based LLMs to enhance the capabilities of handling extensive context windows across various development stages with a holistic taxonomy that categorizes these methodologies targeting different module designs in Transformer. Then, we explore evaluation necessities specific to long-text tasks and some optimization toolkits that integrate many tools to augment LLMs' efficiency and efficacy. We further identify key challenges with corresponding future directions. In addition, our repository ensures that readers stay updated with the latest research in this dynamic field. As LLMs continue to evolve rapidly, we sincerely hope our survey serves as a valuable resource for researchers seeking to harness their power in building powerful long-context LLMs, ultimately advancing the pursuit of the era of AGI.
    

\begin{acks}
We would like to express our gratitude to Hao Gao, Zenan Li, Linyun Liu, etc for their helpful discussions and feedback during the early stages of this paper. Then we acknowledge the generous support from the Baidu AI Cloud Group (ACG), and genuine appreciation to Dou Shen, the Executive Vice President and Head of ACG, for the great idea and gracious invitation to the first session of Baidu ACG Summer Camp. This opportunity has been instrumental in shaping our research and providing valuable experiences.
\end{acks}

    \bibliographystyle{ACM-Reference-Format}
\bibliography{ref-acm}
    
    \newpage
\appendix


\section{Datasets}\label{appendix: datasets}

\begin{table}[htbp]
\centering
\caption{Basic information about existing datasets specific for various NLP tasks with long-text inputs.}
\label{tab: datasets}
\resizebox{\textwidth}{!}{%

\begin{tabular}{|c|c|c|cccccccccc|ccc|cc|c|c|c|}
\hline
\multirow{2}{*}{\textbf{Dataset}} & \multirow{2}{*}{\textbf{Language}} & \multirow{2}{*}{\textbf{\begin{tabular}[c]{@{}c@{}}Task \\ Amount\end{tabular}}} & \multicolumn{10}{c|}{\textbf{Task Types}} & \multicolumn{3}{c|}{\textbf{\begin{tabular}[c]{@{}c@{}}Lengths \\ (kilo words)\end{tabular}}} & \multicolumn{2}{c|}{\textbf{Quality}} & \multirow{2}{*}{\textbf{Splits}} & \multirow{2}{*}{\textbf{Count}} & \multirow{2}{*}{\textbf{Format}} \\ \cline{4-18}
 &  &  & \multicolumn{1}{c|}{\textit{LM}} & \multicolumn{1}{c|}{\textit{MCQA}} & \multicolumn{1}{c|}{\textit{ExtQA}} & \multicolumn{1}{c|}{\textit{Summ}} & \multicolumn{1}{c|}{\textit{Class}} & \multicolumn{1}{c|}{\textit{Match}} & \multicolumn{1}{c|}{\textit{Math}} & \multicolumn{1}{c|}{\textit{Code}} & \multicolumn{1}{c|}{\textit{OpenW}} & \textit{MT} & \multicolumn{1}{c|}{\textit{Avg}} & \multicolumn{1}{c|}{\textit{Min}} & \textit{Max} & \multicolumn{1}{c|}{\textit{\begin{tabular}[c]{@{}c@{}}Human\\ Labeled\end{tabular}}} & \textit{\begin{tabular}[c]{@{}c@{}}Model\\ Assisted\end{tabular}} &  &  &  \\ \hline
\href{https://github.com/armancohan/long-summarization}{ArXiv + PubMed} & en & 1 & \multicolumn{1}{c|}{\coloredxmark} & \multicolumn{1}{c|}{\coloredxmark} & \multicolumn{1}{c|}{\coloredxmark} & \multicolumn{1}{c|}{\coloredcheckmark} & \multicolumn{1}{c|}{\coloredxmark} & \multicolumn{1}{c|}{\coloredxmark} & \multicolumn{1}{c|}{\coloredxmark} & \multicolumn{1}{c|}{\coloredxmark} & \multicolumn{1}{c|}{\coloredxmark} & \coloredxmark & \multicolumn{1}{c|}{5.2} & \multicolumn{1}{c|}{0} & 157.3 & \multicolumn{1}{c|}{\coloredcheckmark} & \coloredxmark & train/test/val & 322K/13.1K/13.1K & jsonl \\ \hline
\href{https://github.com/evasharma/bigpatent}{BigPatent} & en & 1 & \multicolumn{1}{c|}{\coloredxmark} & \multicolumn{1}{c|}{\coloredxmark} & \multicolumn{1}{c|}{\coloredxmark} & \multicolumn{1}{c|}{\coloredcheckmark} & \multicolumn{1}{c|}{\coloredxmark} & \multicolumn{1}{c|}{\coloredxmark} & \multicolumn{1}{c|}{\coloredxmark} & \multicolumn{1}{c|}{\coloredxmark} & \multicolumn{1}{c|}{\coloredxmark} & \coloredxmark & \multicolumn{1}{c|}{3.2} & \multicolumn{1}{c|}{0.2} & 83.2 & \multicolumn{1}{c|}{\coloredcheckmark} & \coloredxmark & train/test/val & 1.2M/67.1K/67.1K & json \\ \hline
\href{https://huggingface.co/datasets/kmfoda/booksum}{BookSum} & en & 1 & \multicolumn{1}{c|}{\coloredxmark} & \multicolumn{1}{c|}{\coloredxmark} & \multicolumn{1}{c|}{\coloredxmark} & \multicolumn{1}{c|}{\coloredcheckmark} & \multicolumn{1}{c|}{\coloredxmark} & \multicolumn{1}{c|}{\coloredxmark} & \multicolumn{1}{c|}{\coloredxmark} & \multicolumn{1}{c|}{\coloredxmark} & \multicolumn{1}{c|}{\coloredxmark} & \coloredxmark & \multicolumn{1}{c|}{4.5} & \multicolumn{1}{c|}{0.04} & 115.8 & \multicolumn{1}{c|}{\coloredcheckmark} & \coloredxmark & train/test/val & 9.6K/1.4K/1.5K & csv \\ \hline
\href{https://github.com/china-ai-law-challenge/CAIL2019/tree/master/scm}{CAIL2019-SCM} & zh & 1 & \multicolumn{1}{c|}{\coloredxmark} & \multicolumn{1}{c|}{\coloredxmark} & \multicolumn{1}{c|}{\coloredxmark} & \multicolumn{1}{c|}{\coloredxmark} & \multicolumn{1}{c|}{\coloredxmark} & \multicolumn{1}{c|}{\coloredcheckmark} & \multicolumn{1}{c|}{\coloredxmark} & \multicolumn{1}{c|}{\coloredxmark} & \multicolumn{1}{c|}{\coloredxmark} & \coloredxmark & \multicolumn{1}{c|}{2.0} & \multicolumn{1}{c|}{1.8} & 2.6 & \multicolumn{1}{c|}{\coloredcheckmark} & \coloredxmark & train/test/val & 5.1K/1.5K/1.5K & jsonl \\ \hline
\href{https://github.com/simengsun/chapterbreak}{ChapterBreak} & en & 1 & \multicolumn{1}{c|}{\coloredxmark} & \multicolumn{1}{c|}{\coloredcheckmark} & \multicolumn{1}{c|}{\coloredxmark} & \multicolumn{1}{c|}{\coloredxmark} & \multicolumn{1}{c|}{\coloredxmark} & \multicolumn{1}{c|}{\coloredxmark} & \multicolumn{1}{c|}{\coloredxmark} & \multicolumn{1}{c|}{\coloredxmark} & \multicolumn{1}{c|}{\coloredxmark} & \coloredxmark & \multicolumn{1}{c|}{25.4} & \multicolumn{1}{c|}{2.3} & 405.8 & \multicolumn{1}{c|}{\coloredcheckmark} & \coloredxmark & train & 9.6K & json \\ \hline
\href{https://github.com/theamrzaki/text_summurization_abstractive_methods}{CNN/DailyMail} & en & 1 & \multicolumn{1}{c|}{\coloredxmark} & \multicolumn{1}{c|}{\coloredxmark} & \multicolumn{1}{c|}{\coloredxmark} & \multicolumn{1}{c|}{\coloredcheckmark} & \multicolumn{1}{c|}{\coloredxmark} & \multicolumn{1}{c|}{\coloredxmark} & \multicolumn{1}{c|}{\coloredxmark} & \multicolumn{1}{c|}{\coloredxmark} & \multicolumn{1}{c|}{\coloredxmark} & \coloredxmark & \multicolumn{1}{c|}{0.8} & \multicolumn{1}{c|}{0} & 2.9 & \multicolumn{1}{c|}{\coloredcheckmark} & \coloredxmark & test & 312K & txt \\ \hline
\href{https://github.com/stanfordnlp/contract-nli-bert}{ContractNLI} & en & 1 & \multicolumn{1}{c|}{\coloredxmark} & \multicolumn{1}{c|}{\coloredxmark} & \multicolumn{1}{c|}{\coloredxmark} & \multicolumn{1}{c|}{\coloredxmark} & \multicolumn{1}{c|}{\coloredxmark} & \multicolumn{1}{c|}{\coloredxmark} & \multicolumn{1}{c|}{\coloredcheckmark} & \multicolumn{1}{c|}{\coloredxmark} & \multicolumn{1}{c|}{\coloredxmark} & \coloredxmark & \multicolumn{1}{c|}{2.0} & \multicolumn{1}{c|}{0.5} & 8.7 & \multicolumn{1}{c|}{\coloredcheckmark} & \coloredxmark & train/test/dev & 423/123/61 & json \\ \hline
\href{https://github.com/PaddlePaddle/Research/tree/master/NLP/ACL2022-DuLeMon}{DuLeMon} & zh & 1 & \multicolumn{1}{c|}{\coloredcheckmark} & \multicolumn{1}{c|}{\coloredxmark} & \multicolumn{1}{c|}{\coloredxmark} & \multicolumn{1}{c|}{\coloredxmark} & \multicolumn{1}{c|}{\coloredxmark} & \multicolumn{1}{c|}{\coloredxmark} & \multicolumn{1}{c|}{\coloredxmark} & \multicolumn{1}{c|}{\coloredxmark} & \multicolumn{1}{c|}{\coloredxmark} & \coloredxmark & \multicolumn{1}{c|}{0.6} & \multicolumn{1}{c|}{0.3} & 1.4 & \multicolumn{1}{c|}{\coloredcheckmark} & \coloredxmark & train/test/dev & 25.4K/1.1K/1.1K & jsonl \\ \hline
\href{https://huggingface.co/datasets/huynguyendayrui/ecthr}{ECtHR} & en & 1 & \multicolumn{1}{c|}{\coloredxmark} & \multicolumn{1}{c|}{\coloredxmark} & \multicolumn{1}{c|}{\coloredxmark} & \multicolumn{1}{c|}{\coloredxmark} & \multicolumn{1}{c|}{\coloredcheckmark} & \multicolumn{1}{c|}{\coloredxmark} & \multicolumn{1}{c|}{\coloredxmark} & \multicolumn{1}{c|}{\coloredxmark} & \multicolumn{1}{c|}{\coloredxmark} & \coloredxmark & \multicolumn{1}{c|}{2.2} & \multicolumn{1}{c|}{0.01} & 51.3 & \multicolumn{1}{c|}{\coloredcheckmark} & \coloredxmark & train/test/dev & 7.3K/3K/1.3K & jsonl \\ \hline
\href{https://github.com/luyang-huang96/LongDocSum}{GovReport} & en & 1 & \multicolumn{1}{c|}{\coloredxmark} & \multicolumn{1}{c|}{\coloredxmark} & \multicolumn{1}{c|}{\coloredxmark} & \multicolumn{1}{c|}{\coloredcheckmark} & \multicolumn{1}{c|}{\coloredxmark} & \multicolumn{1}{c|}{\coloredxmark} & \multicolumn{1}{c|}{\coloredxmark} & \multicolumn{1}{c|}{\coloredxmark} & \multicolumn{1}{c|}{\coloredxmark} & \coloredxmark & \multicolumn{1}{c|}{43.5} & \multicolumn{1}{c|}{0.2} & 1386.2 & \multicolumn{1}{c|}{\coloredcheckmark} & \coloredxmark & test & 19.4K & json \\ \hline
\href{https://huggingface.co/datasets/hotpot_qa}{HotpotQA} & en & 1 & \multicolumn{1}{c|}{\coloredxmark} & \multicolumn{1}{c|}{\coloredxmark} & \multicolumn{1}{c|}{\coloredcheckmark} & \multicolumn{1}{c|}{\coloredxmark} & \multicolumn{1}{c|}{\coloredxmark} & \multicolumn{1}{c|}{\coloredxmark} & \multicolumn{1}{c|}{\coloredxmark} & \multicolumn{1}{c|}{\coloredxmark} & \multicolumn{1}{c|}{\coloredxmark} & \coloredxmark & \multicolumn{1}{c|}{0.9} & \multicolumn{1}{c|}{0.01} & 2.0 & \multicolumn{1}{c|}{\coloredcheckmark} & \coloredxmark & train/dev & 90K/14.8K & json \\ \hline
\href{https://github.com/OpenBMB/InfiniteBench}{InfiniteBench} & en/zh & 12 & \multicolumn{1}{c|}{\coloredxmark} & \multicolumn{1}{c|}{\coloredcheckmark} & \multicolumn{1}{c|}{\coloredcheckmark} & \multicolumn{1}{c|}{\coloredcheckmark} & \multicolumn{1}{c|}{\coloredxmark} & \multicolumn{1}{c|}{\coloredxmark} & \multicolumn{1}{c|}{\coloredcheckmark} & \multicolumn{1}{c|}{\coloredcheckmark} & \multicolumn{1}{c|}{\coloredxmark} & \coloredxmark & \multicolumn{1}{c|}{71.1} & \multicolumn{1}{c|}{0.1} & 560.3 & \multicolumn{1}{c|}{\coloredcheckmark} & \coloredxmark & test & 3.9K & jsonl \\ \hline
\href{https://huggingface.co/datasets/microsoft/LCC_python}{LCC-Python} & py & 1 & \multicolumn{1}{c|}{\coloredxmark} & \multicolumn{1}{c|}{\coloredxmark} & \multicolumn{1}{c|}{\coloredxmark} & \multicolumn{1}{c|}{\coloredxmark} & \multicolumn{1}{c|}{\coloredxmark} & \multicolumn{1}{c|}{\coloredxmark} & \multicolumn{1}{c|}{\coloredxmark} & \multicolumn{1}{c|}{\coloredcheckmark} & \multicolumn{1}{c|}{\coloredxmark} & \coloredxmark & \multicolumn{1}{c|}{1.4} & \multicolumn{1}{c|}{0.2} & 23.3 & \multicolumn{1}{c|}{\coloredcheckmark} & \coloredxmark & train/test/val & 100K/10K/10K & parquet \\ \hline
\href{https://github.com/OpenLMLab/LEval}{LEval} & en & 20 & \multicolumn{1}{c|}{\coloredxmark} & \multicolumn{1}{c|}{\coloredcheckmark} & \multicolumn{1}{c|}{\coloredcheckmark} & \multicolumn{1}{c|}{\coloredcheckmark} & \multicolumn{1}{c|}{\coloredxmark} & \multicolumn{1}{c|}{\coloredxmark} & \multicolumn{1}{c|}{\coloredcheckmark} & \multicolumn{1}{c|}{\coloredcheckmark} & \multicolumn{1}{c|}{\coloredcheckmark} & \coloredxmark & \multicolumn{1}{c|}{9.2} & \multicolumn{1}{c|}{2.0} & 137.5 & \multicolumn{1}{c|}{\coloredcheckmark} & \coloredcheckmark & test & 537 & jsonl \\ \hline
\href{https://huggingface.co/datasets/Yukang/LongAlpaca-12k}{LongAlpaca} & en & 1 & \multicolumn{1}{c|}{\coloredcheckmark} & \multicolumn{1}{c|}{\coloredxmark} & \multicolumn{1}{c|}{\coloredcheckmark} & \multicolumn{1}{c|}{\coloredcheckmark} & \multicolumn{1}{c|}{\coloredxmark} & \multicolumn{1}{c|}{\coloredxmark} & \multicolumn{1}{c|}{\coloredxmark} & \multicolumn{1}{c|}{\coloredxmark} & \multicolumn{1}{c|}{\coloredxmark} & \coloredxmark & \multicolumn{1}{c|}{6.7} & \multicolumn{1}{c|}{0} & 32.7 & \multicolumn{1}{c|}{\coloredcheckmark} & \coloredxmark & train & 12K & json \\ \hline
\href{https://github.com/THUDM/LongBench}{LongBench} & en/zh & 21 & \multicolumn{1}{c|}{\coloredxmark} & \multicolumn{1}{c|}{\coloredxmark} & \multicolumn{1}{c|}{\coloredcheckmark} & \multicolumn{1}{c|}{\coloredcheckmark} & \multicolumn{1}{c|}{\coloredcheckmark} & \multicolumn{1}{c|}{\coloredcheckmark} & \multicolumn{1}{c|}{\coloredxmark} & \multicolumn{1}{c|}{\coloredcheckmark} & \multicolumn{1}{c|}{\coloredxmark} & \coloredxmark & \multicolumn{1}{c|}{7.2} & \multicolumn{1}{c|}{0.1} & 44.2 & \multicolumn{1}{c|}{\coloredcheckmark} & \coloredcheckmark & test & 8.4K & jsonl \\ \hline
\href{https://huggingface.co/datasets/abacusai/LongChat-Lines}{LongChat-Lines} & en & 1 & \multicolumn{1}{c|}{\coloredcheckmark} & \multicolumn{1}{c|}{\coloredxmark} & \multicolumn{1}{c|}{\coloredxmark} & \multicolumn{1}{c|}{\coloredxmark} & \multicolumn{1}{c|}{\coloredxmark} & \multicolumn{1}{c|}{\coloredxmark} & \multicolumn{1}{c|}{\coloredxmark} & \multicolumn{1}{c|}{\coloredxmark} & \multicolumn{1}{c|}{\coloredcheckmark} & \coloredxmark & \multicolumn{1}{c|}{2.6} & \multicolumn{1}{c|}{0.6} & 5.6 & \multicolumn{1}{c|}{\coloredcheckmark} & \coloredxmark & test & 700 & parquet \\ \hline
\href{https://github.com/thu-coai/LOT-LongLM}{LOT} & zh & 4 & \multicolumn{1}{c|}{\coloredxmark} & \multicolumn{1}{c|}{\coloredxmark} & \multicolumn{1}{c|}{\coloredxmark} & \multicolumn{1}{c|}{\coloredxmark} & \multicolumn{1}{c|}{\coloredxmark} & \multicolumn{1}{c|}{\coloredcheckmark} & \multicolumn{1}{c|}{\coloredxmark} & \multicolumn{1}{c|}{\coloredxmark} & \multicolumn{1}{c|}{\coloredcheckmark} & \coloredxmark & \multicolumn{1}{c|}{0.2} & \multicolumn{1}{c|}{0.06} & 0.5 & \multicolumn{1}{c|}{\coloredcheckmark} & \coloredxmark & train/test/dev & 35.2K/2.4K/1.8K & jsonl \\ \hline
\href{https://github.com/google-research/long-range-arena}{LRA - AAN} & en & 1 & \multicolumn{1}{c|}{\coloredxmark} & \multicolumn{1}{c|}{\coloredxmark} & \multicolumn{1}{c|}{\coloredxmark} & \multicolumn{1}{c|}{\coloredxmark} & \multicolumn{1}{c|}{\coloredcheckmark} & \multicolumn{1}{c|}{\coloredcheckmark} & \multicolumn{1}{c|}{\coloredxmark} & \multicolumn{1}{c|}{\coloredxmark} & \multicolumn{1}{c|}{\coloredxmark} & \coloredxmark & \multicolumn{1}{c|}{4.7} & \multicolumn{1}{c|}{0.02} & 55.5 & \multicolumn{1}{c|}{\coloredcheckmark} & \coloredxmark & train/test/dev & 147K/17.4K/18K & tsv \\ \hline
\href{https://github.com/google-research/long-range-arena}{LRA - ListOps} & en & 1 & \multicolumn{1}{c|}{\coloredxmark} & \multicolumn{1}{c|}{\coloredxmark} & \multicolumn{1}{c|}{\coloredxmark} & \multicolumn{1}{c|}{\coloredxmark} & \multicolumn{1}{c|}{\coloredcheckmark} & \multicolumn{1}{c|}{\coloredxmark} & \multicolumn{1}{c|}{\coloredxmark} & \multicolumn{1}{c|}{\coloredxmark} & \multicolumn{1}{c|}{\coloredxmark} & \coloredxmark & \multicolumn{1}{c|}{3} & \multicolumn{1}{c|}{0.01} & 5.9 & \multicolumn{1}{c|}{\coloredcheckmark} & \coloredxmark & train/test/dev & 96K/2K/2K & tsv \\ \hline
\href{https://github.com/ghomashudson/muld}{MuLD} & en & 6 & \multicolumn{1}{c|}{\coloredxmark} & \multicolumn{1}{c|}{\coloredxmark} & \multicolumn{1}{c|}{\coloredcheckmark} & \multicolumn{1}{c|}{\coloredcheckmark} & \multicolumn{1}{c|}{\coloredcheckmark} & \multicolumn{1}{c|}{\coloredxmark} & \multicolumn{1}{c|}{\coloredxmark} & \multicolumn{1}{c|}{\coloredxmark} & \multicolumn{1}{c|}{\coloredxmark} & \coloredcheckmark & \multicolumn{1}{c|}{27.7} & \multicolumn{1}{c|}{0} & 359.1 & \multicolumn{1}{c|}{\coloredcheckmark} & \coloredxmark & train/test/val & 155.9K/14.4K/11.6K & jsonl \\ \hline
\href{https://github.com/Alex-Fabbri/Multi-News}{MultiNews} & en & 1 & \multicolumn{1}{c|}{\coloredxmark} & \multicolumn{1}{c|}{\coloredxmark} & \multicolumn{1}{c|}{\coloredxmark} & \multicolumn{1}{c|}{\coloredcheckmark} & \multicolumn{1}{c|}{\coloredxmark} & \multicolumn{1}{c|}{\coloredxmark} & \multicolumn{1}{c|}{\coloredxmark} & \multicolumn{1}{c|}{\coloredxmark} & \multicolumn{1}{c|}{\coloredxmark} & \coloredxmark & \multicolumn{1}{c|}{2.1} & \multicolumn{1}{c|}{0.1} & 464.2 & \multicolumn{1}{c|}{\coloredcheckmark} & \coloredxmark & train/test/val & 45.0K/5.6K/5.6K & txt \\ \hline
\href{https://huggingface.co/datasets/nayohan/multi_session_chat}{Multi-Session Chat} & en & 1 & \multicolumn{1}{c|}{\coloredcheckmark} & \multicolumn{1}{c|}{\coloredxmark} & \multicolumn{1}{c|}{\coloredxmark} & \multicolumn{1}{c|}{\coloredxmark} & \multicolumn{1}{c|}{\coloredxmark} & \multicolumn{1}{c|}{\coloredxmark} & \multicolumn{1}{c|}{\coloredxmark} & \multicolumn{1}{c|}{\coloredxmark} & \multicolumn{1}{c|}{\coloredcheckmark} & \coloredxmark & \multicolumn{1}{c|}{0.3} & \multicolumn{1}{c|}{0.1} & 1.2 & \multicolumn{1}{c|}{\coloredcheckmark} & \coloredxmark & train/test/val & 17.9K/2.5K/3K & parquet \\ \hline
\href{https://huggingface.co/datasets/natural_questions}{Nature Questions} & en & 1 & \multicolumn{1}{c|}{\coloredxmark} & \multicolumn{1}{c|}{\coloredcheckmark} & \multicolumn{1}{c|}{\coloredxmark} & \multicolumn{1}{c|}{\coloredxmark} & \multicolumn{1}{c|}{\coloredxmark} & \multicolumn{1}{c|}{\coloredxmark} & \multicolumn{1}{c|}{\coloredxmark} & \multicolumn{1}{c|}{\coloredxmark} & \multicolumn{1}{c|}{\coloredxmark} & \coloredxmark & \multicolumn{1}{c|}{9.8} & \multicolumn{1}{c|}{0.2} & 169.3 & \multicolumn{1}{c|}{\coloredcheckmark} & \coloredxmark & train/dev & 307K/7.8K & json \\ \hline
\href{http://qwone.com/~jason/20Newsgroups/}{NewsGroups} & en & 1 & \multicolumn{1}{c|}{\coloredxmark} & \multicolumn{1}{c|}{\coloredxmark} & \multicolumn{1}{c|}{\coloredxmark} & \multicolumn{1}{c|}{\coloredxmark} & \multicolumn{1}{c|}{\coloredcheckmark} & \multicolumn{1}{c|}{\coloredxmark} & \multicolumn{1}{c|}{\coloredxmark} & \multicolumn{1}{c|}{\coloredxmark} & \multicolumn{1}{c|}{\coloredxmark} & \coloredxmark & \multicolumn{1}{c|}{0.3} & \multicolumn{1}{c|}{0} & 11.8 & \multicolumn{1}{c|}{\coloredcheckmark} & \coloredxmark & test & 20K & txt \\ \hline
\href{https://github.com/lil-lab/newsroom}{NewsRoom} & en & 1 & \multicolumn{1}{c|}{\coloredxmark} & \multicolumn{1}{c|}{\coloredxmark} & \multicolumn{1}{c|}{\coloredxmark} & \multicolumn{1}{c|}{\coloredcheckmark} & \multicolumn{1}{c|}{\coloredxmark} & \multicolumn{1}{c|}{\coloredxmark} & \multicolumn{1}{c|}{\coloredxmark} & \multicolumn{1}{c|}{\coloredxmark} & \multicolumn{1}{c|}{\coloredxmark} & \coloredxmark & \multicolumn{1}{c|}{0.7} & \multicolumn{1}{c|}{0} & 178.5 & \multicolumn{1}{c|}{\coloredcheckmark} & \coloredxmark & train/test/dev & 995.0K/108.9K/108.8K & jsonl \\ \hline
\href{https://huggingface.co/datasets/openchat/openchat_sharegpt4_dataset}{OpenChat-ShareGPT4-Clean} & en & 1 & \multicolumn{1}{c|}{\coloredcheckmark} & \multicolumn{1}{c|}{\coloredxmark} & \multicolumn{1}{c|}{\coloredxmark} & \multicolumn{1}{c|}{\coloredxmark} & \multicolumn{1}{c|}{\coloredxmark} & \multicolumn{1}{c|}{\coloredxmark} & \multicolumn{1}{c|}{\coloredxmark} & \multicolumn{1}{c|}{\coloredxmark} & \multicolumn{1}{c|}{\coloredcheckmark} & \coloredxmark & \multicolumn{1}{c|}{1.6} & \multicolumn{1}{c|}{0} & 152.8 & \multicolumn{1}{c|}{\coloredcheckmark} & \coloredcheckmark & train & 80.2K & json \\ \hline
\href{https://huggingface.co/datasets/hoskinson-center/proofnet}{ProofNet} & en & 1 & \multicolumn{1}{c|}{\coloredxmark} & \multicolumn{1}{c|}{\coloredxmark} & \multicolumn{1}{c|}{\coloredxmark} & \multicolumn{1}{c|}{\coloredxmark} & \multicolumn{1}{c|}{\coloredxmark} & \multicolumn{1}{c|}{\coloredxmark} & \multicolumn{1}{c|}{\coloredcheckmark} & \multicolumn{1}{c|}{\coloredxmark} & \multicolumn{1}{c|}{\coloredxmark} & \coloredxmark & \multicolumn{1}{c|}{0.2} & \multicolumn{1}{c|}{0.05} & 0.7 & \multicolumn{1}{c|}{\coloredcheckmark} & \coloredxmark & test/val & 186/185 & jsonl \\ \hline
\href{https://github.com/Yale-LILY/QMSum}{QMSum} & en & 1 & \multicolumn{1}{c|}{\coloredxmark} & \multicolumn{1}{c|}{\coloredxmark} & \multicolumn{1}{c|}{\coloredxmark} & \multicolumn{1}{c|}{\coloredcheckmark} & \multicolumn{1}{c|}{\coloredxmark} & \multicolumn{1}{c|}{\coloredxmark} & \multicolumn{1}{c|}{\coloredxmark} & \multicolumn{1}{c|}{\coloredxmark} & \multicolumn{1}{c|}{\coloredxmark} & \coloredxmark & \multicolumn{1}{c|}{10.8} & \multicolumn{1}{c|}{1.7} & 26.8 & \multicolumn{1}{c|}{\coloredcheckmark} & \coloredxmark & train/test/val & 162/35/35 & jsonl \\ \hline
\href{https://github.com/tau-nlp/scrolls}{SCROLLS} & en & 7 & \multicolumn{1}{c|}{\coloredxmark} & \multicolumn{1}{c|}{\coloredcheckmark} & \multicolumn{1}{c|}{\coloredcheckmark} & \multicolumn{1}{c|}{\coloredcheckmark} & \multicolumn{1}{c|}{\coloredxmark} & \multicolumn{1}{c|}{\coloredxmark} & \multicolumn{1}{c|}{\coloredcheckmark} & \multicolumn{1}{c|}{\coloredxmark} & \multicolumn{1}{c|}{\coloredxmark} & \coloredxmark & \multicolumn{1}{c|}{33.0} & \multicolumn{1}{c|}{0.2} & 356.1 & \multicolumn{1}{c|}{\coloredcheckmark} & \coloredxmark & train/test/val & 89.7K/17.5K/12.3K & jsonl \\ \hline
\href{https://huggingface.co/datasets/squad}{SQuAD} & en & 1 & \multicolumn{1}{c|}{\coloredxmark} & \multicolumn{1}{c|}{\coloredxmark} & \multicolumn{1}{c|}{\coloredcheckmark} & \multicolumn{1}{c|}{\coloredxmark} & \multicolumn{1}{c|}{\coloredxmark} & \multicolumn{1}{c|}{\coloredxmark} & \multicolumn{1}{c|}{\coloredxmark} & \multicolumn{1}{c|}{\coloredxmark} & \multicolumn{1}{c|}{\coloredxmark} & \coloredxmark & \multicolumn{1}{c|}{0.1} & \multicolumn{1}{c|}{0.02} & 0.7 & \multicolumn{1}{c|}{\coloredcheckmark} & \coloredxmark & train/val & 87.6K/10.6K & parquet \\ \hline
\href{https://github.com/mingdachen/SummScreen}{SummScreen} & en & 1 & \multicolumn{1}{c|}{\coloredxmark} & \multicolumn{1}{c|}{\coloredxmark} & \multicolumn{1}{c|}{\coloredxmark} & \multicolumn{1}{c|}{\coloredcheckmark} & \multicolumn{1}{c|}{\coloredxmark} & \multicolumn{1}{c|}{\coloredxmark} & \multicolumn{1}{c|}{\coloredxmark} & \multicolumn{1}{c|}{\coloredxmark} & \multicolumn{1}{c|}{\coloredxmark} & \coloredxmark & \multicolumn{1}{c|}{7.3} & \multicolumn{1}{c|}{1.6} & 24.0 & \multicolumn{1}{c|}{\coloredcheckmark} & \coloredxmark & train/test/dev & 22.6K/2.1K/2.1K & jsonl \\ \hline
\href{https://huggingface.co/datasets/google/Synthetic-Persona-Chat}{Synthetic-Persona-Chat} & en & 1 & \multicolumn{1}{c|}{\coloredcheckmark} & \multicolumn{1}{c|}{\coloredxmark} & \multicolumn{1}{c|}{\coloredxmark} & \multicolumn{1}{c|}{\coloredxmark} & \multicolumn{1}{c|}{\coloredxmark} & \multicolumn{1}{c|}{\coloredxmark} & \multicolumn{1}{c|}{\coloredxmark} & \multicolumn{1}{c|}{\coloredxmark} & \multicolumn{1}{c|}{\coloredcheckmark} & \coloredxmark & \multicolumn{1}{c|}{0.4} & \multicolumn{1}{c|}{0.05} & 0.8 & \multicolumn{1}{c|}{\coloredcheckmark} & \coloredcheckmark & train/test/val & 8.9K/968/1K & csv \\ \hline
\href{https://github.com/ShannonAI/ChineseBert/tree/main/tasks/THUCNew}{THUCnews} & zh & 1 & \multicolumn{1}{c|}{\coloredxmark} & \multicolumn{1}{c|}{\coloredxmark} & \multicolumn{1}{c|}{\coloredxmark} & \multicolumn{1}{c|}{\coloredxmark} & \multicolumn{1}{c|}{\coloredcheckmark} & \multicolumn{1}{c|}{\coloredxmark} & \multicolumn{1}{c|}{\coloredxmark} & \multicolumn{1}{c|}{\coloredxmark} & \multicolumn{1}{c|}{\coloredxmark} & \coloredxmark & \multicolumn{1}{c|}{0.9} & \multicolumn{1}{c|}{0} & 79.5 & \multicolumn{1}{c|}{\coloredcheckmark} & \coloredxmark & test & 836K & txt \\ \hline
\href{https://huggingface.co/datasets/stingning/ultrachat}{UltraChat} & en & 1 & \multicolumn{1}{c|}{\coloredcheckmark} & \multicolumn{1}{c|}{\coloredxmark} & \multicolumn{1}{c|}{\coloredxmark} & \multicolumn{1}{c|}{\coloredxmark} & \multicolumn{1}{c|}{\coloredxmark} & \multicolumn{1}{c|}{\coloredxmark} & \multicolumn{1}{c|}{\coloredxmark} & \multicolumn{1}{c|}{\coloredxmark} & \multicolumn{1}{c|}{\coloredcheckmark} & \coloredxmark & \multicolumn{1}{c|}{1.0} & \multicolumn{1}{c|}{0.03} & 3.6 & \multicolumn{1}{c|}{\coloredcheckmark} & \coloredcheckmark & train & 1.4M & jsonl \\ \hline
\href{https://huggingface.co/datasets/abacusai/WikiQA-Altered_Numeric_QA}{WikiQA-AlteredNumericQA} & en & 1 & \multicolumn{1}{c|}{\coloredxmark} & \multicolumn{1}{c|}{\coloredxmark} & \multicolumn{1}{c|}{\coloredcheckmark} & \multicolumn{1}{c|}{\coloredxmark} & \multicolumn{1}{c|}{\coloredxmark} & \multicolumn{1}{c|}{\coloredxmark} & \multicolumn{1}{c|}{\coloredxmark} & \multicolumn{1}{c|}{\coloredxmark} & \multicolumn{1}{c|}{\coloredxmark} & \coloredxmark & \multicolumn{1}{c|}{4.0} & \multicolumn{1}{c|}{0.8} & 11.2 & \multicolumn{1}{c|}{\coloredcheckmark} & \coloredxmark & test & 1.8K & parquet \\ \hline
\href{https://huggingface.co/datasets/abacusai/WikiQA-Free_Form_QA}{WikiQA-FreeFormQA} & en & 1 & \multicolumn{1}{c|}{\coloredxmark} & \multicolumn{1}{c|}{\coloredxmark} & \multicolumn{1}{c|}{\coloredcheckmark} & \multicolumn{1}{c|}{\coloredxmark} & \multicolumn{1}{c|}{\coloredxmark} & \multicolumn{1}{c|}{\coloredxmark} & \multicolumn{1}{c|}{\coloredxmark} & \multicolumn{1}{c|}{\coloredxmark} & \multicolumn{1}{c|}{\coloredxmark} & \coloredxmark & \multicolumn{1}{c|}{3.8} & \multicolumn{1}{c|}{0.6} & 11.5 & \multicolumn{1}{c|}{\coloredcheckmark} & \coloredxmark & test & 2.4K & parquet \\ \hline
\href{https://huggingface.co/datasets/wmt14}{WMT14 EN-CS} & en/cs & 1 & \multicolumn{1}{c|}{\coloredxmark} & \multicolumn{1}{c|}{\coloredxmark} & \multicolumn{1}{c|}{\coloredxmark} & \multicolumn{1}{c|}{\coloredxmark} & \multicolumn{1}{c|}{\coloredxmark} & \multicolumn{1}{c|}{\coloredxmark} & \multicolumn{1}{c|}{\coloredxmark} & \multicolumn{1}{c|}{\coloredxmark} & \multicolumn{1}{c|}{\coloredxmark} & \coloredcheckmark & \multicolumn{1}{c|}{0.04} & \multicolumn{1}{c|}{0} & 3.6 & \multicolumn{1}{c|}{\coloredcheckmark} & \coloredxmark & train/test/cal & 1M/3K/3K & sgm \\ \hline
\href{https://huggingface.co/datasets/EdinburghNLP/xsum}{XSum} & en & 1 & \multicolumn{1}{c|}{\coloredxmark} & \multicolumn{1}{c|}{\coloredxmark} & \multicolumn{1}{c|}{\coloredxmark} & \multicolumn{1}{c|}{\coloredcheckmark} & \multicolumn{1}{c|}{\coloredxmark} & \multicolumn{1}{c|}{\coloredxmark} & \multicolumn{1}{c|}{\coloredxmark} & \multicolumn{1}{c|}{\coloredxmark} & \multicolumn{1}{c|}{\coloredxmark} & \coloredxmark & \multicolumn{1}{c|}{0.4} & \multicolumn{1}{c|}{0} & 29.2 & \multicolumn{1}{c|}{\coloredcheckmark} & \coloredxmark & train/test/val & 204K/11.3K/11.3K & summary \\ \hline
\end{tabular}%

}

\vspace{1ex} 
\fontsize{6}{10}\selectfont
\begin{minipage}{\textwidth}
\textbf{Note 1}: We sort datasets at each row in the alphabetical order and use slash "/" to separate the multiple contents in any single cell.

\textbf{Note 2}: The presence of common dirty data may result in extremely short samples, thus many datasets in the table containing samples with a minimum length approaching zero.
\end{minipage}

\end{table}

\noindent The meta information regarding the columns in Tab.~\ref{tab: datasets} is as follows:
\begin{itemize}
    \item \textbf{Language}: Language information is represented using abbreviations such as \textit{en} for English, \textit{zh} for Chinese, and \textit{py} for Python.
    
    \item \textbf{Task Amount and Types}: We categorize the common NLP tasks into ten types, including language modeling (\textit{LM}), multi-choice question-answering (\textit{MCQA}), extractive question-answering with information retrieval (\textit{ExtQA}), document summarization (\textit{Summ}), text classification (\textit{Class}), text-pair matching (\textit{Match}), math problem solving and reasoning (\textit{Math}), code tasks (\textit{Code}), open-ended writing (\textit{OpenW}), and machine translation (\textit{MT}).
    
    \item \textbf{Lengths}: Average (\textit{avg}), minimum (\textit{min}), and maximum (\textit{max}) sample lengths are provided in kilo "words"\footnote{In the context of our study, "words" are approximately considered to be separated by spaces in English and code, while individual Chinese characters are treated as words, to avoid the inconsistency by different tokenizers.} for each dataset, where "words" are defined based on sample content\footnote{For example, if one typical sample has the prompt template like \textit{"Read this $\{$context$\}$, and answer the question below: $\{$question$\}$"}, we will calculate the number of words in both context and question part, ignoring the fixed remaining part in the template.}.
    
    \item \textbf{Quality}: Quality assessment is simply based on two dimensions: \textit{Human Labeled} (labels generated by humans) and \textit{Model Assisted} (prompts or labels generated by off-the-shelf LLMs), since the lack of quantitative oracles. 
    
    \item \textbf{Splits}: This indicates dataset partitioning, including conventional triple-split formats like \textit{train/test/val}, a single \textit{test} split for evaluation, a single \textit{train} split for training/finetuning, etc.
    
    \item \textbf{Count}: Provides statistics on the number of samples for each split (one unit "K"/"M" equals 1,000/1,000,000 samples).
    
    \item \textbf{Format}: Tags the file format of samples, including \textit{jsonl}, \textit{json}, \textit{csv}, \textit{txt}, \textit{tsv}, \textit{parquet}, and more.
\end{itemize}


\section{Metrics} \label{appendix: metrics}

\begin{table}[htbp]
\centering
\caption{Some common metrics adopted for evaluation on each specific NLP task type as depicted in Appendix.~\ref{appendix: datasets}.}
\label{tab: metrics}
\resizebox{\textwidth}{!}{%

\begin{tabular}{|c|ccccccccc|}
\hline
\multirow{2}{*}{\textbf{Task Types}} & \multicolumn{9}{c|}{\textbf{Metric Types}} \\ \cline{2-10} 
 & \multicolumn{1}{c|}{\textit{CE/PPL}} & \multicolumn{1}{c|}{\textit{BPC/BPW}} & \multicolumn{1}{c|}{\textit{Acc/F1}} & \multicolumn{1}{c|}{\textit{EM}} & \multicolumn{1}{c|}{\textit{ROUGE-1/-2/-L}} & \multicolumn{1}{c|}{\textit{BLEU/METEOR/TER}} & \multicolumn{1}{c|}{\textit{EntMent}} & \multicolumn{1}{c|}{\textit{Pass@k}} & \textit{Human/Model Judge} \\ \hline
LM & \multicolumn{1}{c|}{\coloredcheckmark} & \multicolumn{1}{c|}{\coloredcheckmark} & \multicolumn{1}{c|}{\coloredcheckmark} & \multicolumn{1}{c|}{\coloredxmark} & \multicolumn{1}{c|}{\coloredxmark} & \multicolumn{1}{c|}{\coloredxmark} & \multicolumn{1}{c|}{\coloredxmark} & \multicolumn{1}{c|}{\coloredxmark} & \coloredcheckmark \\ \hline
MCQA & \multicolumn{1}{c|}{\coloredxmark} & \multicolumn{1}{c|}{\coloredxmark} & \multicolumn{1}{c|}{\coloredcheckmark} & \multicolumn{1}{c|}{\coloredxmark} & \multicolumn{1}{c|}{\coloredxmark} & \multicolumn{1}{c|}{\coloredxmark} & \multicolumn{1}{c|}{\coloredxmark} & \multicolumn{1}{c|}{\coloredcheckmark} & \coloredxmark \\ \hline
ExtQA & \multicolumn{1}{c|}{\coloredxmark} & \multicolumn{1}{c|}{\coloredxmark} & \multicolumn{1}{c|}{\coloredcheckmark} & \multicolumn{1}{c|}{\coloredcheckmark} & \multicolumn{1}{c|}{\coloredcheckmark} & \multicolumn{1}{c|}{\coloredcheckmark} & \multicolumn{1}{c|}{\coloredcheckmark} & \multicolumn{1}{c|}{\coloredxmark} & \coloredcheckmark \\ \hline
Summ & \multicolumn{1}{c|}{\coloredxmark} & \multicolumn{1}{c|}{\coloredxmark} & \multicolumn{1}{c|}{\coloredxmark} & \multicolumn{1}{c|}{\coloredxmark} & \multicolumn{1}{c|}{\coloredcheckmark} & \multicolumn{1}{c|}{\coloredcheckmark} & \multicolumn{1}{c|}{\coloredcheckmark} & \multicolumn{1}{c|}{\coloredxmark} & \coloredcheckmark \\ \hline
Class & \multicolumn{1}{c|}{\coloredxmark} & \multicolumn{1}{c|}{\coloredxmark} & \multicolumn{1}{c|}{\coloredcheckmark} & \multicolumn{1}{c|}{\coloredxmark} & \multicolumn{1}{c|}{\coloredxmark} & \multicolumn{1}{c|}{\coloredxmark} & \multicolumn{1}{c|}{\coloredxmark} & \multicolumn{1}{c|}{\coloredxmark} & \coloredxmark \\ \hline
Match & \multicolumn{1}{c|}{\coloredxmark} & \multicolumn{1}{c|}{\coloredxmark} & \multicolumn{1}{c|}{\coloredcheckmark} & \multicolumn{1}{c|}{\coloredxmark} & \multicolumn{1}{c|}{\coloredxmark} & \multicolumn{1}{c|}{\coloredxmark} & \multicolumn{1}{c|}{\coloredxmark} & \multicolumn{1}{c|}{\coloredcheckmark} & \coloredxmark \\ \hline
Math & \multicolumn{1}{c|}{\coloredxmark} & \multicolumn{1}{c|}{\coloredxmark} & \multicolumn{1}{c|}{\coloredcheckmark} & \multicolumn{1}{c|}{\coloredcheckmark} & \multicolumn{1}{c|}{\coloredxmark} & \multicolumn{1}{c|}{\coloredxmark} & \multicolumn{1}{c|}{\coloredxmark} & \multicolumn{1}{c|}{\coloredcheckmark} & \coloredcheckmark \\ \hline
Code & \multicolumn{1}{c|}{\coloredxmark} & \multicolumn{1}{c|}{\coloredxmark} & \multicolumn{1}{c|}{\coloredcheckmark} & \multicolumn{1}{c|}{\coloredcheckmark} & \multicolumn{1}{c|}{\coloredcheckmark} & \multicolumn{1}{c|}{\coloredcheckmark} & \multicolumn{1}{c|}{\coloredcheckmark} & \multicolumn{1}{c|}{\coloredcheckmark} & \coloredcheckmark \\ \hline
OpenW & \multicolumn{1}{c|}{\coloredxmark} & \multicolumn{1}{c|}{\coloredxmark} & \multicolumn{1}{c|}{\coloredcheckmark} & \multicolumn{1}{c|}{\coloredcheckmark} & \multicolumn{1}{c|}{\coloredcheckmark} & \multicolumn{1}{c|}{\coloredcheckmark} & \multicolumn{1}{c|}{\coloredcheckmark} & \multicolumn{1}{c|}{\coloredcheckmark} & \coloredcheckmark \\ \hline
MT & \multicolumn{1}{c|}{\coloredxmark} & \multicolumn{1}{c|}{\coloredxmark} & \multicolumn{1}{c|}{\coloredxmark} & \multicolumn{1}{c|}{\coloredxmark} & \multicolumn{1}{c|}{\coloredxmark} & \multicolumn{1}{c|}{\coloredcheckmark} & \multicolumn{1}{c|}{\coloredcheckmark} & \multicolumn{1}{c|}{\coloredcheckmark} & \coloredcheckmark \\ \hline
\end{tabular}%

}

\vspace{1ex} 
\fontsize{6}{10}\selectfont
\begin{minipage}{\textwidth}
\textbf{Note}: The $\;$ \coloredxmark $\;$ in the table does not imply that a specific metric cannot be applied to a task. Rather, it suggests that the metric might be less commonly used or that there could be more suitable alternatives.
\end{minipage}

\end{table}

\smallskip
\noindent We provide a concise introduction to these metrics:
\begin{itemize}
    \item \textbf{CE/PPL (Cross-Entropy/Perplexity)}: CE quantifies the $KL$ divergence between predicted distributions and the true distribution from the training corpus. PPL measures how well a language model predicts a sequence, simply formalized as $\exp (loss)$, where $loss$ denotes the cross-entropy loss for the test set.
    
    \item \textbf{BPC/BPW (Bits per Character/Bits per Word)}: They measure the average number of bits required to encode characters or words, i.e. assess the efficiency of a model to compress text, simply calculated by $\mathrm{avg}_{\mathrm{T}}(loss)$, where $T$ is the number of characters or words respectively.
    
    \item \textbf{Acc/F1 (Accuracy/F1 Score)}: Accuracy measures correct predictions in tasks with objective answers like classification and MCQA, while F1 balances precision and recall, as a more robust accuracy score.
    
    \item \textbf{EM (Exact Matching)}: Evaluates exact sequence matches, crucial for tasks like code completion.
    
    \item \textbf{ROUGE-1/-2/-L}~\cite{ganesan2018rouge}: Assess text similarity using n-grams overlapping, typically setting n=1 (unigram), 2 (bigram), and L (longest). They are widely used in tasks that EM may fail, such as summarization.
    
    \item \textbf{BLEU/METEOR/TER}~\cite{agarwal2008meteor}: These metrics are specific in machine translation tasks. BLEU measures the overlap of generated and reference translations based on n-grams. METEOR evaluates translation quality by considering various linguistic factors. TER quantifies the edit distance between the generated and reference translations.
    
    \item \textbf{EntMent (Entity Mention)}~\cite{dai2018entity}: Evaluates coverage and correctness of important entities mentioned in the generated output text, especially for summarization.
    
    \item \textbf{Pass@k}: Evaluates if the generated answer ranks within the top-k provided answers, commonly used in code generation and some math tasks with multiple possible solutions.
    
    \item \textbf{Human/Model Judge}: Involves human or power models like GPT-4 to score text quality based on fluency, coherence, and other subjective criteria suitable for tasks like story generation.
\end{itemize}


\section{Baselines} \label{appendix: baselines}

\begin{table}[htbp]
\centering
\caption{Basic information for some long-context models widely-used as baselines.}
\label{tab: baselines}
\resizebox{\textwidth}{!}{%

\begin{tabular}{|c|c|c|c|c|c|c|c|c|c|}
\hline
\textbf{Model} & \textbf{Open Source} & \textbf{Base} & \textbf{Main Usage} & \textbf{Main Lang} & \textbf{$\bm{L_{max}}$ (k)} & \textbf{Param Size (B)} & \textbf{Mem Occ (GB)} & \textbf{Disk Occ (GB)} & \textbf{Links} \\ \hline
Anima-7B-100k & \coloredcheckmark & Llama2 & chat & zh & 100 & 6.7 & 12.6 & 12.6 & \href{https://huggingface.co/lyogavin/Anima-7B-100K}{hf} | \href{https://github.com/lyogavin/Anima/tree/main/anima_100k}{github} \\ \hline
ChatGLM2-6B-32k & \coloredcheckmark & GLM & chat & zh & 32 & 6.2 & 11.7 & 11.6 & \href{https://huggingface.co/THUDM/chatglm2-6b-32k}{hf} | \href{https://github.com/THUDM/ChatGLM2-6B}{github} \\ \hline
ChatGLM3-6B-32k & \coloredcheckmark & GLM & chat & zh & 32 & 6.2 & 11.7 & 11.6 & \href{https://huggingface.co/THUDM/chatglm3-6b}{hf} | \href{https://github.com/THUDM/ChatGLM3}{github} \\ \hline
Chinese-Alpaca2-7B-16k & \coloredcheckmark & Llama2 & instruct & zh & 16 & 6.9 & 25.9 & 12.9 & \href{https://huggingface.co/hfl/chinese-alpaca-2-7b-16k}{hf} | \href{https://github.com/ymcui/Chinese-LLaMA-Alpaca-2/}{github} \\ \hline
Chinese-Llama2-7B-16k & \coloredcheckmark & Llama2 & chat & zh & 16 & 6.9 & 26.3 & 12.9 & \href{https://huggingface.co/hfl/chinese-llama-2-7b-16k}{hf} | \href{https://github.com/ymcui/Chinese-LLaMA-Alpaca-2/}{github} \\ \hline
Chinese-Mixtral & \coloredcheckmark & Mixtral & chat & zh & 32 & 46.7 & 175.0 & 87.0 & \href{https://huggingface.co/hfl/chinese-mixtral}{hf} | \href{https://github.com/ymcui/Chinese-Mixtral}{github} \\ \hline
Chinese-Mixtral-Instruct & \coloredcheckmark & Mixtral & instruct & zh & 32 & 46.7 & 175.0 & 87.0 & \href{https://huggingface.co/hfl/chinese-mixtral-instruct}{hf} | \href{https://github.com/ymcui/Chinese-Mixtral}{github} \\ \hline
Claude2 & \coloredxmark & Claude & chat & en & 100 & ? & ? & ? & \href{https://claude.ai/onboarding}{acc} | \href{https://www.anthropic.com/news/claude-2}{home} \\ \hline
CodeLlama-7B & \coloredcheckmark & Llama2 & code & py & 16 & 6.7 & 25.6 & 12.6 & \href{https://huggingface.co/codellama/CodeLlama-7b-hf}{hf} | \href{https://huggingface.co/codellama}{home} | \href{https://arxiv.org/pdf/2308.12950.pdf}{paper} \\ \hline
CodeLlama-13B & \coloredcheckmark & Llama2 & code & py & 16 & 13.0 & 49.1 & 24.2 & \href{https://huggingface.co/codellama/CodeLlama-13b-hf}{hf} | \href{https://huggingface.co/codellama}{home} | \href{https://arxiv.org/pdf/2308.12950.pdf}{paper} \\ \hline
CodeLlama-34B & \coloredcheckmark & Llama2 & code & py & 16 & 33.7 & 126.5 & 62.9 & \href{https://huggingface.co/codellama/CodeLlama-34b-hf}{hf} | \href{https://huggingface.co/codellama}{home} | \href{https://arxiv.org/pdf/2308.12950.pdf}{paper} \\ \hline
Giraffe-13B-32k-v3 & \coloredcheckmark & Llama2 & instruct & en & 32 & 13.0 & 48.6 & 24.2 & \href{https://huggingface.co/abacusai/Giraffe-13b-32k-v3}{hf} | \href{https://github.com/abacusai/long-context}{github} | \href{https://arxiv.org/pdf/2308.10882.pdf}{paper} \\ \hline
Giraffe-v2-70B-32k & \coloredcheckmark & Llama2 & instruct & en & 32 & 69.0 & 227.4 & 128.5 & \href{https://huggingface.co/abacusai/Giraffe-v2-70b-32k}{hf} | \href{https://github.com/abacusai/long-context}{github} | \href{https://arxiv.org/pdf/2308.10882.pdf}{paper} \\ \hline
GPT3.5-Turbo-16k & \coloredxmark & GPT3 & chat & en & 16 & ? & ? & ? & \href{https://chat.openai.com/auth/login}{acc} | \href{https://openai.com/chatgpt}{home} | \href{https://platform.openai.com/docs/models/gpt-3-5-turbo}{doc} \\ \hline
GPT4 & \coloredxmark & GPT4 & chat & en & 8 & ? & ? & ? & \href{https://chat.openai.com/auth/login}{acc} | \href{https://openai.com/gpt-4}{home} | \href{https://platform.openai.com/docs/models/gpt-4-and-gpt-4-turbo}{doc} \\ \hline
GPT4-32k & \coloredxmark & GPT4 & chat & en & 32 & ? & ? & ? & \href{https://chat.openai.com/auth/login}{acc} | \href{https://openai.com/gpt-4}{home} | \href{https://platform.openai.com/docs/models/gpt-4-and-gpt-4-turbo}{doc} \\ \hline
GPT4-Turbo & \coloredxmark & GPT4 & chat & en & 128 & ? & ? & ? & \href{https://chat.openai.com/auth/login}{acc} | \href{https://openai.com/gpt-4}{home} | \href{https://platform.openai.com/docs/models/gpt-4-and-gpt-4-turbo}{doc} \\ \hline
InternLM-Chat-7B & \coloredcheckmark & Llama2 & chat & en & 200 & 6.7 & 12.6 & 12.6 & \href{https://huggingface.co/internlm/internlm2-chat-7b}{hf} | \href{https://github.com/InternLM/InternLM}{github} \\ \hline
Llama2-7B-32k & \coloredcheckmark & Llama2 & chat & en & 32 & 6.7 & 12.6 & 12.6 & \href{https://huggingface.co/togethercomputer/LLaMA-2-7B-32K}{hf} | \href{https://www.together.ai/}{home} \\ \hline
Llama2-7B-Instruct-32k & \coloredcheckmark & Llama2 & instruct & en & 32 & 6.7 & 12.6 & 12.6 & \href{https://huggingface.co/togethercomputer/Llama-2-7B-32K-Instruct}{hf} | \href{https://www.together.ai/}{home} \\ \hline
LLongMA2-7B-16k-flash & \coloredcheckmark & Llama2 & chat & en & 16 & 6.7 & 12.6 & 12.6 & \href{https://huggingface.co/emozilla/LLongMA-2-7b-16k-flash}{hf} | \href{https://arxiv.org/pdf/2309.00071.pdf}{paper} \\ \hline
LongChat-v1.5-7B-32k & \coloredcheckmark & Llama2 & chat & en & 32 & 6.7 & 12.6 & 12.6 & \href{https://huggingface.co/lmsys/longchat-7b-v1.5-32k}{hf} | \href{https://github.com/DachengLi1/LongChat}{github} | \href{https://lmsys.org/blog/2023-06-29-longchat/}{blog} \\ \hline
Mistral-7B-v0.1 & \coloredcheckmark & Mistral & chat & en & 32 & 7.2 & 28.0 & 13.5 & \href{https://huggingface.co/mistralai/Mistral-7B-v0.1}{hf} | \href{https://arxiv.org/pdf/2310.06825.pdf}{paper} \\ \hline
Mistral-7B-Instruct-v0.2 & \coloredcheckmark & Mistral & instruct & en & 32 & 7.2 & 28.0 & 13.5 & \href{https://huggingface.co/mistralai/Mistral-7B-Instruct-v0.2}{hf} | \href{https://arxiv.org/pdf/2310.06825.pdf}{paper} \\ \hline
Mixtral-8x7B-v0.1 & \coloredcheckmark & Mixtral & chat & en & 32 & 46.7 & 175.0 & 87.0 & \href{https://huggingface.co/mistralai/Mixtral-8x7B-v0.1}{hf} | \href{https://mistral.ai/news/mixtral-of-experts/}{blog} \\ \hline
Mixtral-8x7B-Instruct-v0.1 & \coloredcheckmark & Mixtral & instruct & en & 32 & 46.7 & 175.0 & 87.0 & \href{https://huggingface.co/mistralai/Mixtral-8x7B-Instruct-v0.1}{hf} | \href{https://mistral.ai/news/mixtral-of-experts/}{blog} \\ \hline
MPT-7B-Storywriter & \coloredcheckmark & MPT & gen & en & 65 & 6.6 & 12.4 & 12.4 & \href{https://huggingface.co/mosaicml/mpt-7b-storywriter}{hf} | \href{https://www.mosaicml.com/blog/mpt-7b}{blog} \\ \hline
NeuralChat-7B-v3.1 & \coloredcheckmark & Mistral & chat & en & 32 & 7.2 & 28.0 & 13.5 & \href{https://huggingface.co/Intel/neural-chat-7b-v3-1}{hf} | \href{https://medium.com/intel-analytics-software/the-practice-of-supervised-finetuning-and-direct-preference-optimization-on-habana-gaudi2-a1197d8a3cd3}{blog} \\ \hline
OpenHermes2.5-7B & \coloredcheckmark & Mistral & chat & en & 32 & 7.2 & 28.0 & 13.5 & \href{https://huggingface.co/teknium/OpenHermes-2.5-Mistral-7B}{hf} | \href{https://github.com/sponsors/teknium1}{github} \\ \hline
QWen-7B & \coloredcheckmark & QWen & chat & zh & 32 & 7.7 & 14.4 & 14.4 & \href{https://huggingface.co/Qwen/Qwen-7B}{hf} | \href{https://arxiv.org/pdf/2309.16609.pdf}{paper} \\ \hline
Vicuna-v1.5-7B-16k & \coloredcheckmark & Llama2 & chat & en & 16 & 6.7 & 12.6 & 12.6 & \href{https://huggingface.co/lmsys/vicuna-7b-v1.5-16k}{hf} | \href{https://github.com/lm-sys/FastChat}{github} | \href{https://lmsys.org/blog/2023-03-30-vicuna/}{blog} \\ \hline
WizardCoder-Python-7B-v1.0 & \coloredcheckmark & Llama2 & code & py & 16 & 6.7 & 12.8 & 12.6 & \href{https://huggingface.co/WizardLM/WizardCoder-Python-7B-V1.0}{hf} | \href{https://github.com/nlpxucan/WizardLM}{github} \\ \hline
WizardMath-7B-v1.1 & \coloredcheckmark & Mistral & math & en & 32 & 7.2 & 14.0 & 13.5 & \href{https://huggingface.co/WizardLM/WizardMath-7B-V1.1}{hf} | \href{https://github.com/nlpxucan/WizardLM}{github} \\ \hline
XGen-7B-Instruct-8k & \coloredcheckmark & Llama2 & instruct & en & 8 & 6.7 & 12.6 & 12.6 & \href{https://huggingface.co/Salesforce/xgen-7b-8k-inst}{hf} | \href{https://arxiv.org/pdf/2309.03450.pdf}{paper} \\ \hline
\end{tabular}%

}

\vspace{1ex} 
\fontsize{6}{10}\selectfont
\begin{minipage}{\textwidth}
\textbf{Note}: The rows are basically sort by the model names in the alphabetical order, and we use question mark "?" to indicate unknown information for any cell.
\end{minipage}

\end{table}

\noindent Some meta information about Tab.~\ref{tab: baselines} is interpreted as follows:
\begin{itemize}
    \item \textbf{Open Source}: Indicates whether the model is open-sourced (\coloredcheckmark) or closed-sourced that can be accessible only through official remote API (\coloredxmark).
    \item \textbf{Base}: Specifies the base modeling structure upon which the long-context model is built.
    \item \textbf{Main Usage}: Highlights the primary usage and capability of the model, categorized as \textit{instruct} for instruction-following, \textit{code/math} for code/math-related tasks, \textit{gen} for text generation, and \textit{chat} for general-purpose tasks through chat-like interaction.
    \item \textbf{Main Lang}: Indicates the primary language the model can understand\footnote{Models are typically pretrained on multi-language corpora and may be finetuned for specific languages as needed. So we choose the most suitable one corresponding to its application objectives.}, considering natural language, programming language, etc.
    \item \textbf{$\bm{L_{max}}$}: Represents the maximum context length handled by the model, measured in tokens (one unit "k" equals 1024 tokens).
    \item \textbf{Statistics}: Provides statistics about the model, including the number of parameters, memory footprint, and disk storage. All models are loaded with precision to float16 onto Nvidia GPU(s) without any quantization.
    \item \textbf{Links}: Includes publication links for accessing and learning more about the model, with \textit{hf} indicating the Hugging Face hub for open-sourced models and \textit{acc} representing official access for closed-sourced ones.
\end{itemize}


\section{Toolkits} \label{appendix: toolkits}

\begin{table}[htbp]
\centering
\caption{The toolkits summary for enhancing LLMs efficiency and effectiveness across different stages.}
\label{tab: toolkits}
\resizebox{\textwidth}{!}{%

\begin{tabular}{|c|c|cccc|}
\hline
\multirow{2}{*}{\textbf{Toolkit}} & \multirow{2}{*}{\textbf{Type}} & \multicolumn{4}{c|}{\textbf{Utilities for Stages}} \\ \cline{3-6} 
 &  & \multicolumn{1}{c|}{\textit{Pretraining}} & \multicolumn{1}{c|}{\textit{Finetuning}} & \multicolumn{1}{c|}{\textit{Inference}} & \textit{Application} \\ \hline
\href{https://github.com/huggingface/accelerate}{Accelerate} & library & \multicolumn{2}{c|}{\begin{tabular}[c]{@{}c@{}}Integrate TorchRun, FSDP\\ DeepSpeed, Megatron-LM\\ Local SGD\end{tabular}} & \multicolumn{1}{c|}{} & \begin{tabular}[c]{@{}c@{}}CPU/GPU/ TPUs/Apple Silicon\\ Auto Device Management\end{tabular} \\ \hline
\href{https://github.com/microsoft/autogen}{AutoGen} & framework & \multicolumn{1}{c|}{} & \multicolumn{1}{c|}{} & \multicolumn{1}{c|}{Multi-Config Inference} & Multi-Agent Conversation \\ \hline
\href{https://github.com/TimDettmers/bitsandbytes}{BitsandBytes} & library & \multicolumn{2}{c|}{\begin{tabular}[c]{@{}c@{}}8bit Optimizers\\ 8bit Matrix Multiplication\\ QLoRA\end{tabular}} & \multicolumn{1}{c|}{QLoRA} & \begin{tabular}[c]{@{}c@{}}4bit/8bit \\ Quantization\end{tabular} \\ \hline
\href{https://github.com/hpcaitech/ColossalAI}{Colossal-AI} & library & \multicolumn{1}{c|}{\begin{tabular}[c]{@{}c@{}}Integrate DP, PP, \\ 1D/2D/2.5D/3D TP,\\ ZERO\\ Auto Parallelism\end{tabular}} & \multicolumn{1}{c|}{} & \multicolumn{1}{c|}{} &  \\ \hline
\href{https://github.com/microsoft/DeepSpeed}{DeepSpeed} & framework & \multicolumn{3}{c|}{\begin{tabular}[c]{@{}c@{}}DP, PP, TP,\\ ZERO, Offload,\\ Sparse Attention Kernel\end{tabular}} &  \\ \hline
\href{https://github.com/microsoft/DeepSpeed-MII}{DeepSpeed-MII} & framework & \multicolumn{1}{c|}{} & \multicolumn{1}{c|}{} & \multicolumn{1}{c|}{Dynamic SplitFuse} &  \\ \hline
\href{https://github.com/Dao-AILab/flash-attention}{FlashAttention} & library & \multicolumn{3}{c|}{Kernel-Fused Flash-Attention} &  \\ \hline
\href{https://github.com/huggingface/text-generation-inference}{HuggingFace TGI} & system & \multicolumn{1}{c|}{} & \multicolumn{1}{c|}{} & \multicolumn{1}{c|}{\begin{tabular}[c]{@{}c@{}}TP\\ Optimized Architectures\\ Continuous Batching\\ Quantization\end{tabular}} &  \\ \hline
\href{https://github.com/langchain-ai/langchain}{LangChain} & framework & \multicolumn{1}{c|}{} & \multicolumn{1}{c|}{} & \multicolumn{1}{c|}{} & \begin{tabular}[c]{@{}c@{}}Prompt Management\\ Memory Management\\ Agent Management\end{tabular} \\ \hline
\href{https://github.com/chatchat-space/Langchain-Chatchat}{LangChain-Chatchat} & framework & \multicolumn{1}{c|}{} & \multicolumn{1}{c|}{} & \multicolumn{1}{c|}{} & \begin{tabular}[c]{@{}c@{}}Integrate Langchain\\ RAG\end{tabular} \\ \hline
\href{https://github.com/hiyouga/LLaMA-Factory}{Llama-Factory} & library & \multicolumn{1}{c|}{} & \multicolumn{1}{c|}{\begin{tabular}[c]{@{}c@{}}Integrate LoRA, QLoRA, \\ PPO, DPO,  \\ Reward Modeling\end{tabular}} & \multicolumn{1}{c|}{} &  \\ \hline
\href{https://github.com/NVIDIA/Megatron-LM}{Megatron-LM} & framework & \multicolumn{1}{c|}{\begin{tabular}[c]{@{}c@{}}DP, PP, SP, ZERO\\ Activation Checkpointing\end{tabular}} & \multicolumn{1}{c|}{} & \multicolumn{1}{c|}{} &  \\ \hline
\href{https://www.usenix.org/system/files/osdi22-yu.pdf}{Orca} & system & \multicolumn{1}{c|}{} & \multicolumn{1}{c|}{} & \multicolumn{1}{c|}{\begin{tabular}[c]{@{}c@{}}Iteration-level scheduling\\ Selective Batching\end{tabular}} &  \\ \hline
\href{https://github.com/huggingface/peft}{PEFT} & library & \multicolumn{2}{c|}{\begin{tabular}[c]{@{}c@{}}LoRA, \\ Prefix-Tuning,\\ Prompt-Tuning\end{tabular}} & \multicolumn{1}{c|}{} &  \\ \hline
\href{https://github.com/bigscience-workshop/petals}{Petals} & framework & \multicolumn{1}{c|}{} & \multicolumn{1}{c|}{Multi-Party Distributed Collaboration} & \multicolumn{1}{c|}{} &  \\ \hline
\href{https://github.com/imartinez/privateGPT}{PrivateGPT} & framework & \multicolumn{1}{c|}{} & \multicolumn{1}{c|}{} & \multicolumn{1}{c|}{} & Private RAG \\ \hline
\href{https://pytorch.org/blog/introducing-pytorch-fully-sharded-data-parallel-api/}{Pytorch FSDP} & framework & \multicolumn{2}{c|}{Fully Sharded Data Parallel} & \multicolumn{1}{c|}{} &  \\ \hline
\href{https://pytorch.org/docs/stable/generated/torch.nn.functional.scaled_dot_product_attention.html}{Pytorch SDPA} & function & \multicolumn{3}{c|}{\begin{tabular}[c]{@{}c@{}}Integrate Flash-Attention,\\ Memory-Efficient Attention\end{tabular}} &  \\ \hline
\href{https://github.com/NVIDIA/TensorRT-LLM}{TensorRT-LLM} & library & \multicolumn{1}{c|}{} & \multicolumn{1}{c|}{} & \multicolumn{1}{c|}{\begin{tabular}[c]{@{}c@{}}Python API for \\ TensorRT Engines\\ In-flight Batching\end{tabular}} &  \\ \hline
\href{https://github.com/openai/triton}{Triton} & compiler & \multicolumn{3}{c|}{\begin{tabular}[c]{@{}c@{}}Python API for \\ GPU Kernels\end{tabular}} &  \\ \hline
\href{https://github.com/vllm-project/vllm}{vLLM} & library & \multicolumn{1}{c|}{} & \multicolumn{1}{c|}{} & \multicolumn{1}{c|}{Paged Attention} &  \\ \hline
\href{https://github.com/facebookresearch/xformers}{xFormers} & library & \multicolumn{1}{c|}{Memory-Efficient Attention} & \multicolumn{1}{c|}{} & \multicolumn{1}{c|}{} &  \\ \hline
\end{tabular}%

}

\vspace{1ex} 
\fontsize{6}{10}\selectfont
\begin{minipage}{\textwidth}
\textbf{Note}: We sort the toolkits in each row in the alphabetical order.
\end{minipage}

\end{table}

    


We offer a detailed explanation of Tab.~\ref{tab: toolkits} as follows:

\noindent\textbf{Type}: This column specifies the usage type of each toolkit, including: 

\begin{itemize}
    \item \textit{Library}: Typically found as GitHub projects, these toolkits offer functional implementations of specific tasks or algorithms.
    \item \textit{Framework}: Usually encompass a whole systematic pipeline, consisting of multiple interconnected modules designed to support various aspects of LLMs.
    \item \textit{System}: Offer a complete environment that comes pre-configured with all the necessary components and settings to facilitate the deployment of LLMs.
    \item \textit{Compiler}: Fuse operations and compile them into optimized GPU kernels with specific programming languages to accelerate the execution of LLMs.
\end{itemize}

\noindent\textbf{Stages}: We categorize the whole LLM lifecycle simply into four stages as follows:
\begin{itemize}
    \item \textit{Pretraining}: LLMs undergo unsupervised training on large-scale datasets to learn basic language modeling.
    \item \textit{Finetuning}: LLMs are further trained in a supervised manner on full/partial parameters to adapt them to specific tasks or align them with human values.
    \item \textit{Inference}: Involves feeding prompts into LLMs and generating outputs iteratively using various control strategies.
    \item \textit{Application}: Off-the-shelf and even black-box LLMs are utilized for context-aware tasks, often involving domain-specific local documents.
\end{itemize}

\noindent\textbf{Utilities}: For each toolkit, we highlight diverse utilities with concise keywords to indicate core techniques w.r.t the corresponding stages. Readers can refer to the toolkit links for more detailed information on these utilities if lost on any keyword.

\end{document}